\documentclass[journal,transmag]{IEEEtran}
\ifCLASSINFOpdf
\else
\fi

\usepackage{graphicx}
\newcommand{\method}{Graph Informer}
\newcommand{\citet}[1]{\cite{#1}}
\newcommand{\citep}[1]{\cite{#1}}
\usepackage{subcaption}
\usepackage{amsfonts}       
\usepackage{amsmath}
\usepackage{tensor}
\usepackage{listings}
\usepackage{hyperref}
\usepackage{booktabs}
\usepackage{microtype}
\usepackage{fancyvrb}

\newtheorem{lemma}{Lemma}

\begin{document}
%
\title{Expressive Graph Informer Networks}



\author{\IEEEauthorblockN{Jaak Simm\IEEEauthorrefmark{1}\IEEEauthorrefmark{2},
Adam Arany\IEEEauthorrefmark{1}\IEEEauthorrefmark{2},
Edward~De Brouwer\IEEEauthorrefmark{1}, and
Yves Moreau\IEEEauthorrefmark{1}}
\IEEEauthorblockA{\IEEEauthorrefmark{1}ESAT-STADIUS,
KU Leuven, 
Leuven, 3001, Belgium}
\IEEEauthorblockA{\IEEEauthorrefmark{2}Equal contribution as first authors.}
\thanks{Manuscript received x; revised y. 
Corresponding author: J. Simm (email: jaak.simm@esat.kuleuven.be).

\copyright2020 IEEE. Personal use of this material is permitted.
Permission from IEEE must be obtained for all other uses,
in any current or future media, including reprinting/republishing this material for advertising or promotional purposes, creating new collective works, for resale or redistribution to servers or lists, or reuse of any copyrighted component of this work in other works.
}}





%



\IEEEtitleabstractindextext{%
\begin{abstract}
  
  Applying machine learning to molecules is challenging because of their natural representation as graphs rather than vectors. Several architectures have been recently proposed for deep learning from molecular graphs, but they suffer from information bottlenecks because they only pass information from a graph node to its direct neighbors.
  Here, we introduce a more expressive route-based multi-attention mechanism that incorporates features from routes between node pairs.
  We call the resulting method \method{}.
  A single network layer can therefore attend to nodes several steps away.
  We show empirically that the proposed method compares favorably against existing approaches in two prediction tasks: (1)~13C Nuclear Magnetic Resonance (NMR) spectra, improving the state-of-the-art with an MAE of 1.35 ppm and (2)~predicting drug bioactivity and toxicity.
  Additionally, we develop a variant called injective \method{} that is \emph{provably} as powerful as the Weisfeiler-Lehman test for graph isomorphism.
  Furthermore, we demonstrate that the route information allows the method to be informed about the \emph{nonlocal topology} of the graph and, thus, even go beyond the capabilities of the Weisfeiler-Lehman test.
\end{abstract}

}

\maketitle

\IEEEdisplaynontitleabstractindextext

%
\IEEEpeerreviewmaketitle

\section{Introduction}
\label{sec:intro}

Graphs are used as a natural representation for objects in many domains, such as compounds in computational chemistry or protein--protein interaction networks in systems biology.
Machine learning approaches for graphs fall into two main lines of research: the \emph{spectral} and the \emph{spatial} approach.
The spectral approach relies on the eigenvalue decomposition of the Laplacian of the graph and is well suited for problems involving a single fixed graph structure.

By contrast, \emph{spatial} graph methods work directly on the nodes and edges of the graph.
Convolutional neural networks (CNN) have inspired many spatial methods.
Similarly to a local convolution filter running on the 2D grid of an image, spatial approaches update the hidden vectors of a graph node by aggregating the hidden vectors of its neighbors.
Several spatial graph methods have been proposed, which vary in how to carry out this update.
The most straightforward approach is to sum (or average) the hidden vectors of neighbors, and then transform the results with a linear layer and a nonlinearity (\emph{e.g.}, as proposed in Neural Fingerprints~\citep{duvenaud2015convolutional} and Graph Convolutional Networks~\citep{kipf2016semi}).
Instead of a simple dense feedforward layer, some methods use GRU-based gating~\citep{li2015gated}, edge hidden vectors~\citep{kearnes2016molecular}, and attention to the neighbors~\citep{velivckovic2017graph}.
One of the advantages of spatial methods over spectral methods is that they can be straightforwardly applied to problems involving multiple graphs.

\begin{figure}
  \centering
  \includegraphics[width=\linewidth]{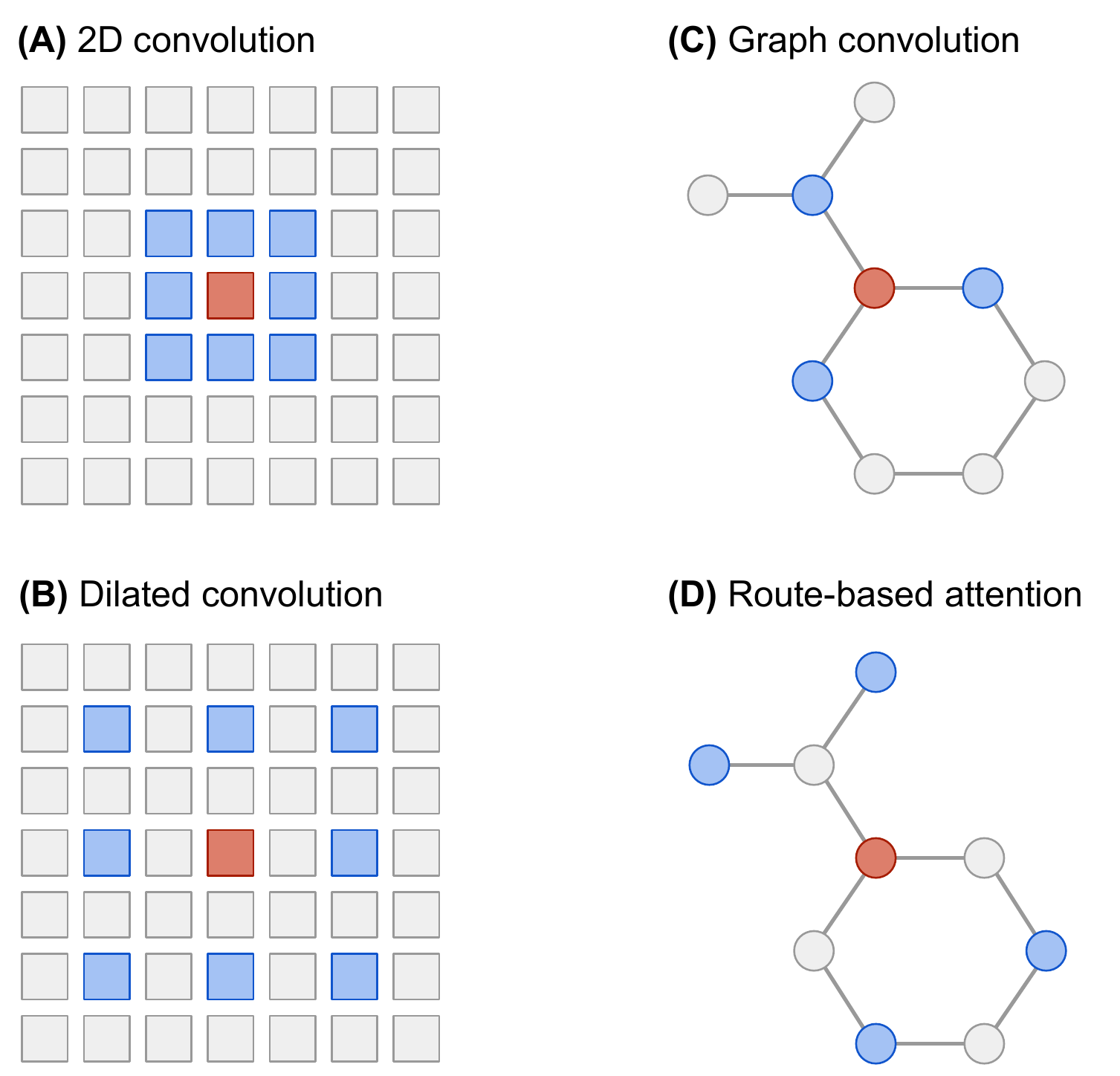}
  \caption{Illustration of how route-based graph attention can mimic dilated filters in graphs. \emph{Red} colored nodes are updated based on the vectors of the \emph{blue} color nodes. (A) 2D convolutional filter, (B) 2D dilated filter, (C) graph convolution, (D) route-based graph attention.}
  \label{fig:route-attention}
\end{figure}

However, the update step of a node in spatial methods only has access to its own neighborhood, which limits the information flow throughout the whole graph. Increasing the accessible neighborhood of each node requires the stacking of several layers.
The dilated filters in CNNs \citep{yu2015MultiScaleContextAggregation} are an example of solving this neighborhood limitation in images, by providing an efficient way to gather information over longer distances (see Figure~\ref{fig:route-attention}, (A) and (B)).

In this work, we propose a method that flexibly aggregates information over longer graph distances in one step, analogously to dilated filters.
Our \emph{\method{}} approach is inspired by the sequence-to-sequence transformer  model~\citep{vaswani2017AttentionAllYou}.
The core contribution of our work is the introduction of \emph{route-based multi-head self-attention} (RouteMHSA), which allows the attention mechanism to also access route information and, thus, base its attention scores both on the features of the nodes and the route between them.
This enables \method{} to gather information from nodes that are not just direct neighbors.
Furthermore, the proposed approach can straightforwardly use edge features, as they can be incorporated into the route features. In the case of chemical compounds, this allows using the annotation of the bonds between the atoms (single, double, aromatic, etc.).
The central idea is illustrated in Figure~\ref{fig:route-attention}.
Additionally, we investigate the expressiveness of \method{}, by describing a variant that is provably as powerful as the Weisfeiler-Lehman (WL) test and showing empirically that it can even go beyond the WL test. 

Our \method{} compares favorably with state-of-the-art graph-based neural networks on 13C NMR spectrum and drug bioactivity prediction tasks, which makes it suitable for both \emph{node-level} and \emph{graph-level} prediction tasks.

\section{Proposed approach}
\subsection{Setup}
In this work, we consider learning problems on graphs where both nodes and edges are labeled.
For a graph $G_i$ with $N_i$ nodes,  we denote its node features by a matrix $X_i \in \mathbb{R}^{N_i \times F_\mathrm{nodes}}$ and the route information by tensor $P_i \in \mathbb{R}^{N_i \times N_i \times F_\mathrm{route}}$, $F_\mathrm{nodes}$ and $F_\mathrm{route}$ are the dimensions of the nodes and route features, respectively and the route information $X_i$ is computed using the adjacency matrix of the graph and the edge labels.
For example, for chemical compound graphs, where nodes are atoms, the vector between a pair of atoms in the route information tensor (\emph{e.g.}, vector $P_i[k,l]$ for atoms $k$ and $l$) can contain information on the type of route that connects the two atoms (rigid, flexible, aromatic) and how far they are from each other (shortest path).

We propose a method that works for both (1) \emph{graph-level} tasks, such as graph classification where the whole graph $G_i$ is assigned a label, and (2) \emph{node-level} tasks where one should predict a value or a label for nodes (or subset of nodes) of the graph.

\subsection{Dot-product self-attention}
Our approach is inspired by the encoder used in the transformer network~\citep{vaswani2017AttentionAllYou}, which proposed dot-product attention for sequence models formulated as
\begin{equation*}
 \mathrm{Attn}(Q, K, V) = \sigma
    \left(
        \frac{1}{\sqrt{d_k}}Q K^\top
    \right) V,
\end{equation*}
where $Q, K \in \mathbb{R}^{N_i \times d_k}$ and $V \in \mathbb{R}^{N_i \times d_v}$ are the query, key, and value matrices and $\sigma$ is the Softmax function, which makes sure that the attention probabilities for each node \emph{sum to one}.
This attention can be used as \emph{self-attention} by projecting the input $d$-dimensional hidden vectors $H \in \mathbb{R}^{N_i \times d}$ into queries, keys and values. $Q$, $K$ and $V$ are computed from the input hidden vectors with corresponding weight matrices $W_Q \in \mathbb{R}^{d_k \times d}, W_K \in \mathbb{R}^{d_k \times d}$ and $W_V \in \mathbb{R}^{d_v \times d}$ (\emph{i.e.}, $Q = W_Q H^T$, $K = W_KH^T$ and $V = W_V H^T$).
Intuitively, the $k$-th row of $Q$ is the query vector of the $k$-th node (attending node), the $l$-th row of $K$ is the key vector of node $l$ (attended node) and the $l$-th row of $V$ is the value vector for node $l$ (value of the attended node).

To encode sequences with dot-product attention, the original transformer network \citep{vaswani2017AttentionAllYou} adds position encodings to the hidden vectors $H$.
In particular, sine waves with different frequencies and phases were used in the original paper to encode the position of the sequence elements.
This allows the network to change attention probabilities based on the positions.

\subsection{Route-based dot-product self-attention}
One aspect to note is that in the case of graphs, there is no analogue for the global position of nodes.
Therefore, we propose a novel \emph{stable relative addressing} mechanism for graphs.
The proposed mechanism adds a new component to the dot-attention that depends the \emph{features of the route} between the two nodes in the graph.
This route component is made up of a query and key part.
For the \emph{route query} $Q_R \in \mathbb{R}^{N_i \times d_r}$ we map the hidden vectors of the nodes ($H$) by linear transformation
\begin{equation*}
    Q_R = W_Q^\mathrm{route} H,
\end{equation*}
where $d_r$ is the dimension of the route query and $W_Q$ is the query weight matrix.
Similarly, for the \emph{route key} $K_R \in \mathbb{R}^{N_i \times N_i \times d_r}$ we map the route features $P$ (for readability we drop the sample index $i$) by linear transformation
\begin{equation*}
K_R = W_K^\mathrm{route} P.
\end{equation*}
The route information allows the attention mechanism to access the topology of the graph.
Note that while $Q_R$ is a matrix of route queries, one for each node of the graph, $K_R$ is a tensor of rank 3, with each slice $K_R[k,l]$ corresponding to the route from node $k$ to node $l$.

The route query and key can be now contracted into a $N_i \times N_i$ matrix of (logit) attention scores, which can be added to the original node-based scores $QK^\top$.
Using the \emph{einsum} convention\footnote{We follow the convention introduced by Numpy and used also in Pytorch.}, let
\begin{align}
  Q_R \otimes K_R = \mathrm{einsum}_{\mathrm{kf,klf}\rightarrow \mathrm{kl}}(Q_R, K_R)
\end{align}
or equivalently
\begin{align}
  (Q_R \otimes K_R)[k,l] = Q_R[k]^\top K_R[k,l].
\end{align}
The intuition behind this tensor product is straight-forward: we compute the scalar product between the query vector $Q_R[k]$ of the node $k$ and the key vector $K_R[k,l]$ of the route from $k$ to $l$.

Now the route-based attention probabilities $A \in [0,1]^{N_i \times N_i}$ are given by
\begin{align}
  A = \sigma
    \left(
        \frac{1}{\sqrt{d_k + d_r}}(Q K^\top + Q_R \otimes K_R)
    \right),
    \label{eq:a-route}
\end{align}
where we also added the size of the route key $d_r$ to the normalization for keeping the values within trainable (non-plateau) ranges of the softmax function.
This enables the network to use both the information on the routes ($Q_R \otimes K_R$), as well as the node hidden vectors ($Q K^\top$), to decide the attention probabilities.

Analogously, we project the route information into the value component $V$ of the dot-product attention.
Firstly, we map the route data $P$ into
\begin{equation*}
V_R = W_V^\mathrm{route} P,    
\end{equation*}
where $V_R \in \mathbb{R}^{N_i \times N_i \times d_v}$ is the tensor of rank 3 with slice $V_R[k,l]$ corresponding to the value vector for the route from node $k$ to node $l$.
The route value tensor $V_R$ can be then weighted according to the attention probabilities $A$ from Eq.~\ref{eq:a-route} giving as the final route-based graph attention:
\begin{align}
  \mathrm{RouteAttn}(Q,K,V,Q_R,K_R,V_R) \nonumber \\
  = AV + \mathrm{einsum}_{\mathrm{kl,klv}\rightarrow\mathrm{kv}}(A, V_R).
  \label{eq:RouteAttn}
\end{align}

\subsection{Locality-constrained attention}
The route-based attention defined in Eq.~\ref{eq:RouteAttn} is general and allows unconstrained access throughout the graph.
In some applications, localized access is preferred and this can be easily forced by adding a masking matrix to the attention scores:
\begin{equation}
    A_\mathrm{LC} = \sigma
    \left(
        \frac{1}{\sqrt{d_k + d_r}}(Q K^\top + Q_R \otimes K_R) + M_\mathrm{route}
    \right),
\label{eq:attention_scores}
\end{equation}
where $M_\mathrm{route} \in \{-\infty,0\}^{N_i \times N_i}$ is 0 for unmasked routes and minus infinity for masked connections (in practice, we use a large negative value).
This allows us, for example, to mask all nodes whose shortest distance from the current node is larger than $3$ or some other number, thereby creating an attention ball around the current node.
Similarly, it is possible to use the mask $M_\mathrm{route}$ to create an attention shell (\emph{i.e.}, attending nodes within a given range of shortest distances).

In our experiments, we only used the attention ball with different values for the radius.

\subsection{Multi-head and mini-batching}
Similarly to the transformer architecture, we use multiple heads.
This means several route-based self-attentions are executed in parallel and their results are concatenated.
This allows the network to attend to several neighbors at the same times.

\paragraph{Pool node} Additionally, we introduce a \emph{pool} node that has a different embedding vector than the graph nodes.
The pool node has no edges to the graph nodes, but is unmasked by $M_\mathrm{route}$, so that the attention mechanism can always attend and read, if required.
This idea has two motivations.
First, it allows the information to be easily shared across the graph.
Second, the pool node can be used as a ``not found'' answer when the attention mechanism does not find the queried node within the graph nodes.

\paragraph{Mini-batching} The proposed algorithm supports mini-batching via a few simple modifications.
To introduce the batch dimension, the input tensors $H$ and $P$ must first be padded with zeros to have the same size, after which they are stacked.
The linear projections $Q,K,V,Q_R,K_R,V_R$ are now indexed by both batch (sample) and head.
We replace all matrix multiplications by batched matrix products and in $\mathrm{einsum}$ we introduce leading batch and head dimensions.
For the attention mechanism, we add an additional mask over the nodes, $M_\mathrm{node}$, with large negative values for non-existing nodes (padded) for each sample:
\begin{align*}
    A_\mathrm{LCB} = \sigma
    \left(
        \frac{1}{\sqrt{d_k + d_r}}(Q K^\top + Q_R \otimes K_R) + M_\mathrm{LCB}
    \right),
\end{align*}
where $M_\mathrm{LCB} = M_\mathrm{route} + M_\mathrm{node}$.

\paragraph{Learnable module} Finally, we use the mini-batched attention probabilities $A_\mathrm{LCB}$ for $\mathrm{RouteAttn}$ from Eq.~\ref{eq:RouteAttn}, which takes in (1) hidden vectors of the nodes $H$, with size $(B, N_\mathrm{nodes}, N_\mathrm{hidden})$, (2) route information $P$, with size $(B, N_\mathrm{nodes}, N_\mathrm{nodes}, F_\mathrm{route})$, and (3) masks $M_\mathrm{node}$ and $M_\mathrm{route}$.
We use multiple heads and concatenate their results, giving us a route-based multi-head self-attention module ($\mathrm{RouteMHSA}$):
\begin{align*}
    \mathrm{RouteMHSA}
    &= \mathrm{Concat}(X_1, \ldots, X_{N_\mathrm{head}})
    \\
    \text{with }
    X_i &= \mathrm{RouteAttn}(
        W_{i,Q} H,
        W_{i,K} H,
        W_{i,V} H,\\
        & \quad\quad
        W_{i,Q}^\mathrm{route} H,
        W_{i,K}^\mathrm{route} P,
        W_{i,V}^\mathrm{route} P),
\end{align*}
where $W$ are the learnable parameters of the module.

The values for hidden size $N_\mathrm{hidden}$, key size $d_k$, value size $d_v$, route key size $d_r$, and the number of heads $N_\mathrm{heads}$ can be chosen independently of each other.
However, in our experiments we set $N_\mathrm{hidden}$ equal to $d_k N_\mathrm{heads}$, with $N_\mathrm{heads} \in \{6, 8\}$.
We also use $d_v = d_k = d_r$.



\subsection{Computational complexity}
The computational complexity of the $\mathrm{RouteMHSA}$ is quadratic with respect to the number of nodes in the graph, because the attention mechanism computes all pairwise attention probabilities.
In practice, this means that the method can scale to graphs of a few hundred, maximally 1,000 nodes.
This is more than enough for handling drug-like compounds, which typically have less than 100 (non-hydrogen) atoms.


\section{Architecture of the network}
\label{sec:architectures}



The architecture of Graph Informer is inspired by transformer networks \citep{vaswani2017AttentionAllYou}, where the output of the multi-head attention is fed through a linear layer and added to its input, then a subsequent layer normalization \citep{ba2016LayerNormalization} is applied. Additionally, a feedforward network (FFN) was applied to each hidden vector separately.
However, in our graph experiments, for both node-level and graph-level tasks, we found this architecture to be quite difficult to train (for more details, see Appendix~\ref{app:gradient-issue}).

\begin{figure}
  \centering
  \includegraphics[width=0.7\linewidth]{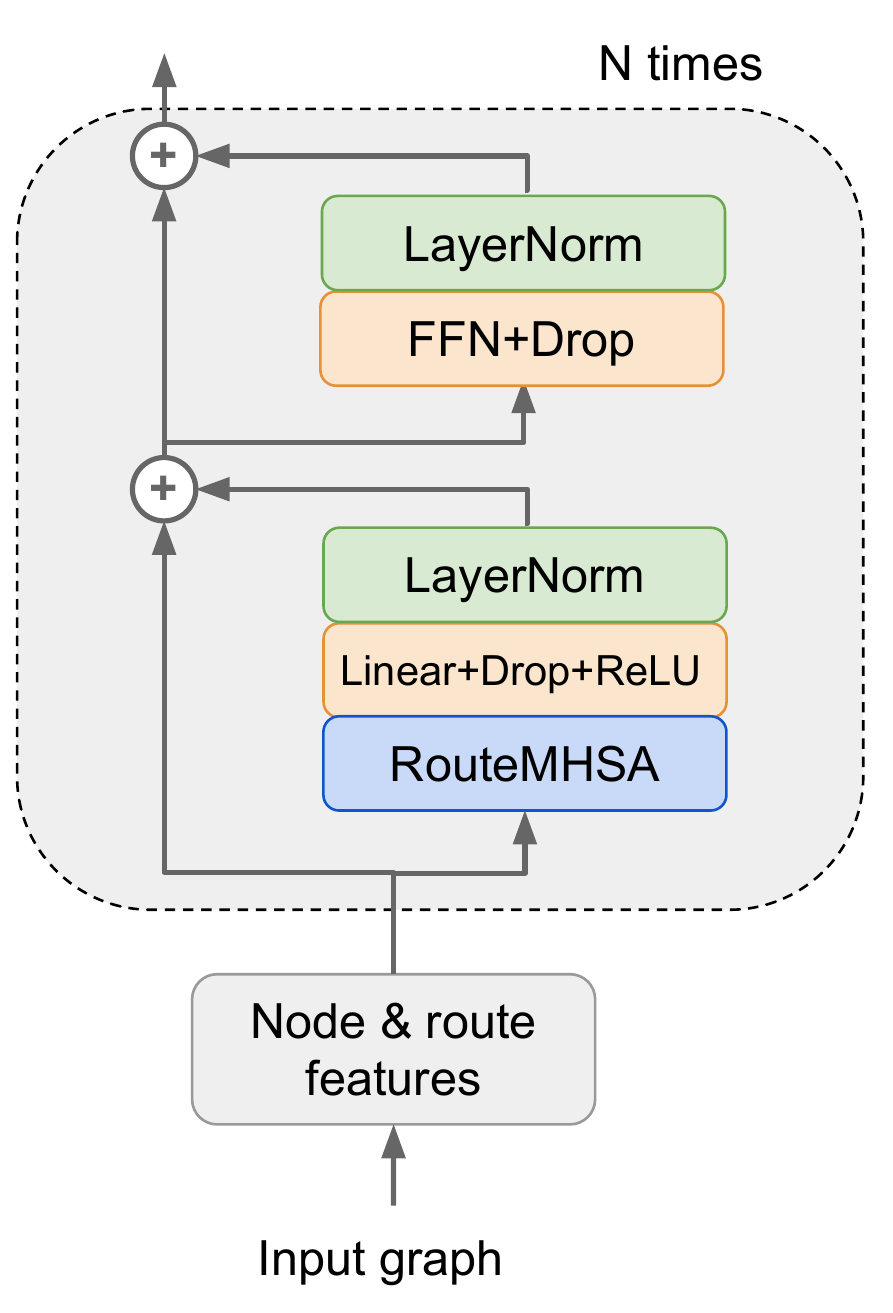}
  \caption{Layer setup for the architectures of \method{}.}
  \label{fig:graph-informer-architectures}
\end{figure}

Instead, we found that creating a residual-style network~\citep{he2016identity} with $\mathrm{LayerNorm}$ solved the gradient flow issue and was easy to train.
The layer for this architecture can be expressed as
\begin{align}
    T  &= H + \mathrm{LayerNorm}(\mathrm{Linear}(\mathrm{RouteMHSA}(H)))
    \\
    H' &= T + \mathrm{LayerNorm}(\mathrm{FFN}(T)),
\end{align}
with $H'$ being the output of the block (\emph{i.e.}, updated hidden vectors). The architecture is depicted in Figure~\ref{fig:graph-informer-architectures}.
\method{} uses both neuron-level and channel-level dropout, with dropout rate 0.1.

\section{Expressiveness of \method{}}
Next, we investigate the general ability of \method{} to distinguish between distinct but similar graphs, also known as the graph isomorphism problem.

\subsection{The Weisfeiler-Lehman test}

We propose a variant of our approach that is provably as powerful as the Weisfeiler-Lehman test of isomorphism of dimension 1 (WL $\mathrm{dim}=1$). Generalizing the notation of \citet{xu2018powerful} to include route information, $P$, we write the update function of our model as
\begin{align*}
    h_v^k = \phi \left( h_v^{k-1},
    \left\{(h_u^{k-1}, P_{v, u}) : u \in \mathcal{N}(v) \right\} \right),
\end{align*}
where  $\left\{ (h_u^{k-1}, P_{v, u}) : u \in \mathcal{N}(v) \right\}$ is the multiset of node vectors, $h_u^{k-1}$, and their route information vectors $P_{v,u}$, which can be attended by node $v$.

Such a network is provably as powerful as the Weisfeiler-Lehman test of isomorphism if the function $\phi(.)$ is injective\footnote{We assume here that the output function is also injective.} with respect to the hidden vectors $h_v^{k-1}$ and $h_u^{k-1}$ arguments, as follows straightforwardly from the theorem by \citet{xu2018powerful}.

$\mathrm{RouteMHSA}$ satisfies this condition with respect to the first argument ($h_v^{k-1}$) because one of the heads can attend to itself. However, to make it \emph{provably} injective with respect to the second argument, we can replace the softmax mapping in Eq.~\ref{eq:attention_scores} by an elementwise sigmoid. We call this version the \emph{Injective} \method{}.

\subsection{Beyond the Weisfeiler-Lehman test}
\begin{figure}[t]
  \centering
    \begin{subfigure}[b]{0.2\textwidth}
        \includegraphics[width=\textwidth]{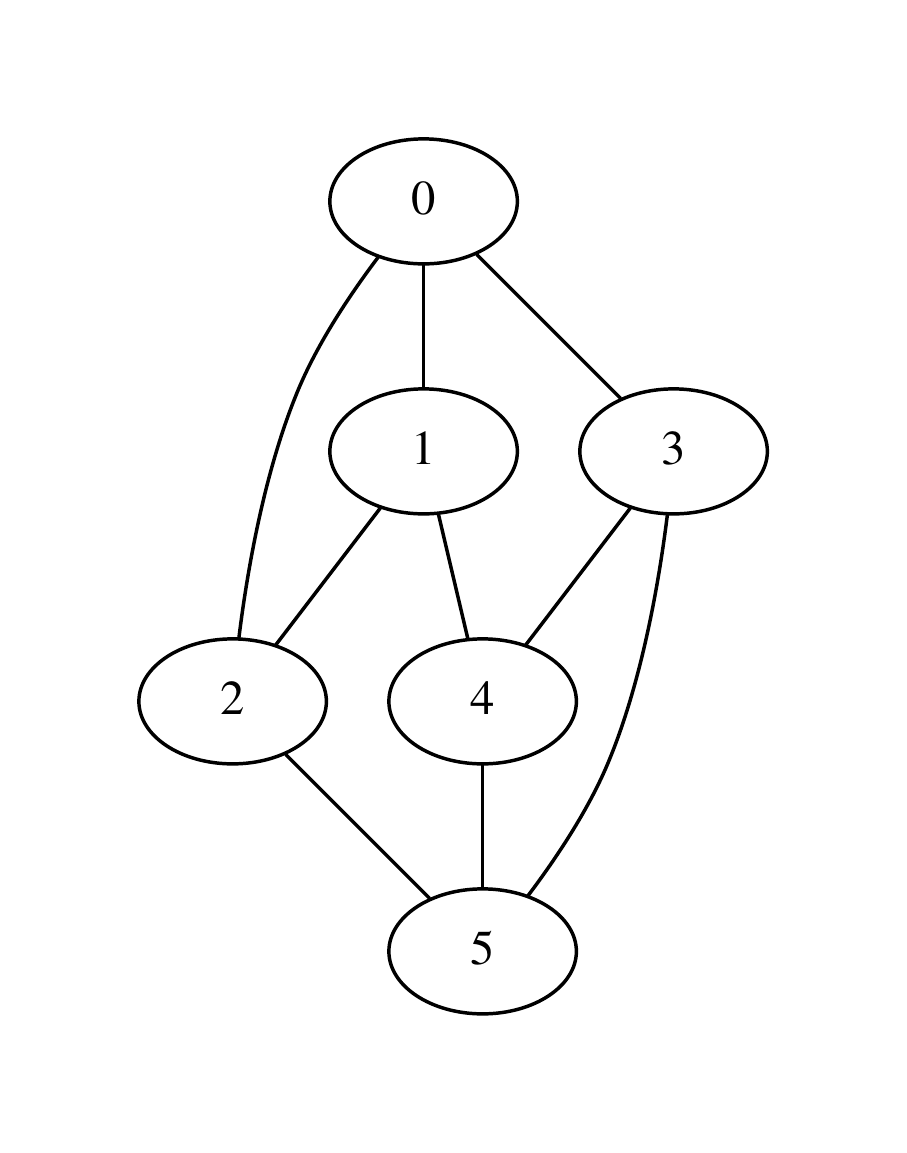}
        \caption{RegN6D3-G1}
        \label{fig:RegN6D3-G1}
    \end{subfigure}
    \begin{subfigure}[b]{0.2\textwidth}
        \includegraphics[width=\textwidth]{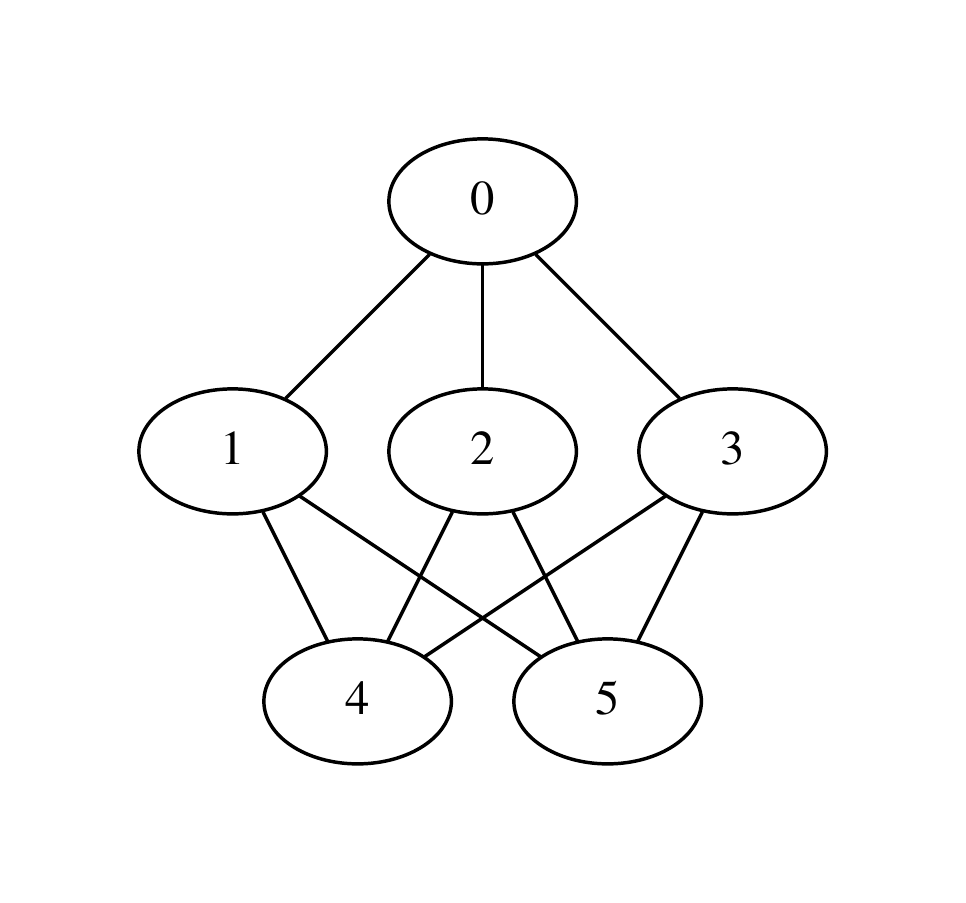}
        \caption{RegN6D3-G2}
        \label{fig:RegN6D3-G2}
    \end{subfigure}
  \caption{All (two) regular 6-node 3-degree graphs. They are distinguished by \method{}.}
  \label{fig:RegN6D3}
\end{figure}

Several graph neural networks, including GIN~\citep{xu2018powerful} and NeuralFP~\citep{duvenaud2015convolutional}, are limited in their expressiveness to the Weisfeiler-Lehman test of $\mathrm{dim}=1$.
This means that if the WL test cannot distinguish two graphs, then also these methods will not be able to separate them.
By contrast, the \method{} can aggregate route information and, therefore, has the capability to go beyond WL $\mathrm{dim}=1$ as the routes between the nodes reveal nonlocal topology.
There are several types of graphs known to be indistinguishable by the WL $\mathrm{dim}=1$ test. For illustrative purposes, we examine the family of regular graphs.
For an example of two non-isomorphic 6-node regular graphs, see Figure~\ref{fig:RegN6D3}.

\subsubsection{Example: regular graphs}
The WL $\mathrm{dim}=1$ test fails to distinguish \emph{any} regular graphs of the same size and degree because the nodes of the regular graph have all the same degree.
In the case of node-level prediction tasks, this also implies that all nodes of a regular graph get the same prediction, while their neighboring topology can be radically different.

We illustrate the capability of the \method{} to distinguish between all regular 8-node 3-degree graphs (RegN8D3).
As a route feature, we use a \emph{histogram} of the lengths of all routes between two nodes up to length $k$.
In other words, for a given node pair $(a,b)$ the route feature is a vector of size $k$, containing the numbers of (possibly cyclic) routes connecting $a$ and $b$ for lengths $1, \ldots k$:
\begin{align}
    P = \mathrm{Stack}(A, A^2, \ldots, A^k),
    \label{eq:route-histogram}
\end{align}
where $A$ is the adjacency matrix and $\mathrm{Stack}$ creates a tensor of size $N \times N \times k$.
Computing the histogram takes $k$ multiplications of the adjacency matrix, which gives complexity $O(N^2 k)$.

There is a total of five RegN8D3 graphs, depicted in Appendix~\ref{app:beyond-wl}.
It should be noted that one of the RegN8D3 graphs corresponds to a chemically well-known compound called \emph{cubane} (G4 in the appendix). While none of them are separable by the WL test or neural networks such as GIN \citep{xu2018powerful}, GAT \citep{velivckovic2017graph}, and NeuralFP \citep{duvenaud2015convolutional}, a randomly initialized and untrained 1-layer Injective \method{} with $k=4$ can separate them all, for details see the Appendix~\ref{app:beyond-wl}.

Additionally, we ran the same network on all regular graphs up to size 10\footnote{Graphs were taken from \url{http://www.mathe2.uni-bayreuth.de/markus/reggraphs.html}}.
As Table~\ref{tbl:regular-graphs} shows, Graph Informer (both injective and ordinary) can distinguish them all.
This contrasts with the WL test and standard aggregating graph neural networks, which are provably unable to separate any of them (for details, see \citet{xu2018powerful}).

We can show that with these features \method{} has the following capability:
\begin{lemma}
  Injective \method{} using route features from Eq.~\ref{eq:route-histogram} with $k=N$ can distinguish any pair of graphs with different spectra.
\end{lemma}
The proof of this lemma follows from the fact that injective \method{} is injective with respect to route features $P$ and that graphs are cospectral (in terms of their adjacency matrix) if and only if $\mathrm{Tr}(A^r)$ are equal for both matrices for all values of $r$ \citep{alzaga2008spectra}.
Notably, our method offers a more detailed view because it can also distinguish cospectral graphs (\emph{e.g.}, the Tesseract (Q4) and Hoffman graphs, which are both 16-node and 4-regular).

\begin{table}[t]
  \caption{Uniquely distinguishing regular graphs using untrained 1-layer \method{} with 4 heads and using a route length histogram of size 4 as route feature between two nodes. None of them are distinguishable by the WL test.}
  \label{tbl:regular-graphs}
  \begin{center}
  \begin{small}
  \begin{sc}
  \begin{tabular}{lrc}
    \toprule
    Graph set   & Separated / Total  & Proportion \\
    \midrule
    Reg N6 D3   &  2 / 2   &  100\% \\
    Reg N7 D4   &  2 / 2   &  100\% \\
    Reg N8 D3   &  5 / 5   &  100\% \\
    Reg N8 D4   &  6 / 6   &  100\% \\
    Reg N8 D5   &  3 / 3   &  100\% \\
    Reg N9 D4   &  16 / 16   &  100\% \\
    Reg N9 D6   &  4  / 4    &  100\% \\
    Reg N10 D3  &  19 / 19   &  100\% \\
    Reg N10 D4  &  59 / 59   &  100\% \\
    Reg N10 D5  &  60 / 60   &  100\% \\
    Reg N10 D6  &  21 / 21   &  100\% \\
    Reg N10 D7  &  5 / 5     &  100\% \\
    Q4 vs Hoffman & 2 / 2    &  100\% \\
    \bottomrule
  \end{tabular}
  \end{sc}
  \end{small}
  \end{center}
\end{table}

We hypothesize that in practice this capability of \method{} gives strong additional expressiveness and makes it difficult to construct counterexamples because the route information allows the method to capture the nonlocal topology of the graph in detail.

\section{Related research}
There are several neural network approaches that can directly work on the graph inputs.
The key component in many of the works is to propagate hidden vectors of the nodes $H$ using the adjacency matrix\footnote{The adjacency matrix here is assumed to include the self-connection (\emph{i.e.}, $A_{ii} = 1$).} $A$, either by summing the neighbor vectors, $A H$, or by averaging them as $\tilde{A} H$, where $\tilde{A} = A (D)^{-1}$ is the normalized adjacency matrix with $D$ being the diagonal matrix of degrees.
Neural Fingerprint~\citep{duvenaud2015convolutional} proposed to use the propagation $AH$, combined with a node-wise linear layer and a nonlinearity to compute fingerprints for chemical compounds.
Based on the degree of the node, Neural Fingerprint uses a different set of linear weights.
Similarly, Graph Convolutional Networks (GCN, \citet{kipf2016semi}) propose to combine $AH$ with a linear layer and a nonlinearity for semi-supervised learning, but by contrast with Neural Fingerprint, they use the same linear layer for all nodes, irrespective of their degree.
Kearnes \emph{et al.} \citet{kearnes2016molecular} proposed the Weave architecture, where they use embedding vectors also for the edges of the graph and the propagation step updates nodes and edges in a coupled way.

In a separate line of research, \citet{scarselli2008graph} proposed Graph Neural Networks (GNN), which also use $AH$ to propagate the hidden vectors between the nodes, but instead of passing the output to the next layer, the same layer is executed until convergence, which is known as the Almeida-Pineda algorithm \citep{almeida1987learning}.
A recent GNN-based approach is called Gated Graph Neural Networks (GGNN) \citep{li2015gated}, where at each iteration the neighborhood information $AH$ is fed into a GRU~\citep{cho14gru}.
An iterative graph network approach, motivated by the Message Passing algorithm in graphical model inference \citep{pearl2014probabilistic} was proposed by \citet{dai2016discriminative}.
Gilmer et al. \citet{gilmer2017neural} generalized the existing approaches to the Neural Message Passing (NeuralMP) algorithm and showed the effectiveness of the approach for predicting simulated quantum properties based on 3D structures of compounds.

Hamilton et al. \citet{hamilton2017inductive} propose a representation learning method, GraphSAGE, that does not require supervised signals and inductively embeds nodes in large graphs.

The method closest to ours is Graph Attention Networks (GAT) \citet{velivckovic2017graph}.
Similarly to us, they proposed to use multi-head attention to transform the hidden vectors of the graph nodes. However, GAT cannot integrate route features and its attention is only limited to the neighbors of every node.
For graph isomorphism testing, this means it fails to separate any regular graphs, similar to WL $\mathrm{dim=1}$.




In contrast to the spatial methods reviewed above, spectral methods use the graph Fourier basis.
One of the first spectral approaches to introduce this concept  in the context of neural networks was SCNN \citep{bruna2013spectral}.
To avoid the costly eigenvalue decomposition,  Chebyshev polynomials were proposed to define localized filters \citet{defferrard2016convolutional}.
Based on this work,  proposed a neural network that combines     spectral methods with metric learning for graph classification was proposed \citet{li2018adaptive}.

For sake of completeness, we note that another loosely related line of research has been working on the problem of transforming an input graph to a target graph and are known as \emph{graph transformers} \citep{bottou1997global,yun2019graph}. Those works usually do not use the attention mechanism used in the transformer network~\citep{vaswani2017AttentionAllYou}. Yet, in a recent communication, also called graph transformers, \citet{li2019graph} proposed a node-level attention mechanism for the same graph-to-graph task. In contrast, our model tackles node- or graph-level classification and regression tasks, and uses route-level attention mechanism.

In graph isomorphism research, as already mentioned, \citep{xu2018powerful} showed that injective update rule and readout are crucial to guarantee expressiveness at the level of WL $\mathrm{dim}=1$.
In a recent publication, \citep{maron2019provably} proposed a graph neural network as powerful as WL $\mathrm{dim}=3$.

In one of our main application areas, Nuclear Magnetic Resonance (NMR) chemical shift prediction (covered in Section~\ref{sec:nmr}), is an important element in structure elucidation.
Recently, \citet{jonas2019deep} proposed an imitation learning approach to solve this NMR inverse problem.

\section{Evaluation}
We compare our method against several baselines on two tasks. We first consider a \emph{node-level} task where, given the graph of the compound, the goal is to predict 13C NMR peaks for all of the carbon atoms, see Section~\ref{sec:nmr}.
For the \emph{graph-level} tasks, we consider the drug--protein activity data set ChEMBL~\citep{bento2014chembl} and compound toxicity prediction from MoleculeNet~\citep{wu2018moleculenet}, see Section~\ref{sec:chembl}.
In all tasks and for all methods, we used the same feature set (see Appendix~\ref{app:features}). For methods that could not use route features but supported edge features, we used the bond type (single, double, triple, aromatic).

Each method was executed using one NVIDIA P100 with 16GB RAM with the data fed using 9 Intel(R) Xeon(R) 6140 Gold cores.
For the implementation of our algorithm, we used Pytorch.

\subsection{Baselines}
Next, we overview the baselines we used for the experiments.
The code for these methods were available online, for the exact URLs, see Appendix~\ref{app:baselines}

\textbf{GGNN}~\citep{li2015gated}:\quad We used the Pytorch implementation of Gated Graph Neural Networks, which was designed for node-level prediction on graphs. We also modified it to work on graph-level tasks by adding the same graph pooling layer used in our network (global average pooling).
For both tasks, we used 5 propagation steps.

\textbf{NeuralFP}~\citep{duvenaud2015convolutional}:\quad  We implemented the Neural Fingerprint in Pytorch. The implementation follows the original method using softmax graph pooling.
We also extended the method to work on node-level tasks by removing the pooling and feeding the last layer hidden vectors to the regression output layer (see Section~\ref{sec:nmr}).

\textbf{GCN}~\citep{kipf2016semi}:\quad For the Graph Convolutional Networks, we adapted the Pytorch implementation to support mini-batch training.
We also adapted the code to support graph classification by incorporating the same graph pooling layer used in our network (global average pooling).

\textbf{Weave}~\citep{kearnes2016molecular}:\quad For the Weave network, we used the Deepchem implementation.
The implementation only supports graph-level predictions.

\textbf{GAT}~\citep{velivckovic2017graph}:\quad We used the Tensorflow implementation of Graph Attention Networks.
The implementation only supports node-level prediction. 

\textbf{NeuralMP}~\citep{gilmer2017neural}: \quad We used the Tensorflow implementation of the neural message passing model where we used the edge network message function (best performing architecture) and added a node-level readout function for the NMR spectrum prediction (same as in all other compared methods).

\textbf{GIN}~\citep{xu2018powerful} \quad We used the Graph Isomorphism Network with the epsilon parameter set to $0$, as suggested for the best performance of this architecture in their paper.

\subsection{Model selection}
All methods used the validation set to choose the optimal hidden size from  $\{96, 144, 192, 384\}$ and dropout from $\{0.0, 0.1\}$ \citep{srivastava2014dropout}. For our method, attention radius from $\{1,2,3,4\}$ and head count from $\{6,8\}$ were also selected.
All methods were trained with the Adam optimizer~\citep{kingma2014adam}. 
The validation set was used for early stopping (\emph{i.e.}, we executed all methods for 100 epochs and picked the best model).
At epochs 40 and 70, the learning rate was decreased by a factor of 0.3.
Finally, the best early stopped model was used to measure the performance on the test set.

\subsection{Node-level task: NMR 13C spectrum prediction}
\label{sec:nmr}
13C NMR is a cost-efficient spectroscopic technique in organic chemistry for molecular structure identification and structure elucidation.
If 13C NMR measurements can be combined with a highly accurate prediction, based on the graph (structure) of the molecule, it is possible to validate the structure of the sample.
It is crucial to have highly accurate predictions, because in many cases several atoms have peaks that are very close to each other.
If such predictions are available, the chemist can avoid running additional expensive and time-consuming experiments for structure validation.


To create the data set, we used 13C NMR spectra from NMRShiftDB2\footnote{Available here: \url{https://nmrshiftdb.nmr.uni-koeln.de/}}. After basic quality filtering (see Appendix~\ref{app:nmr-dataset}) 25,645 molecules remained with peak positions raging from -10 to 250 ppm (parts per million) with mean and standard deviation of 95 and 52 ppm.

\begin{table}[t]
  \caption{Results for NMR 13C spectrum prediction for models with different number of layers. We report the mean and standard deviation for the test MAE and the median size (\#params column) of the best model over the repeats.
  The bold entries are best or statistically not different from the best.}
  \label{tbl:nmr}
  \begin{center}
  \begin{small}
  \begin{sc}
  \begin{tabular}{lcr}
    \toprule
    Method     & \#layers     & Test MAE  \\ 
    \midrule
    Mean model &            & 46.894  \\ 
    GCN      & 2            & 6.284 $\pm$ 0.053 \\ 
    GCN      & 3            & 8.195 $\pm$ 0.385 \\ 
    GIN      & 2            & 5.015 $\pm$ 0.082 \\ 
    GIN      & 3            & 6.275 $\pm$ 0.306 \\ 
    GGNN     & \emph{na}    & 1.726 $\pm$ 0.055 \\ 
    NeuralMP & \emph{na}    & 1.704 $\pm$ 0.025 \\ 
    GAT      & 1            & 2.889 $\pm$ 0.021 \\ 
    GAT      & 2            & 3.148 $\pm$ 0.067 \\ 
    GAT      & 3            & 5.193 $\pm$ 0.196 \\ 
    NeuralFP & 1            & 3.839 $\pm$ 0.039 \\ 
    NeuralFP & 2            & 2.248 $\pm$ 0.027 \\ 
    NeuralFP & 3            & 2.040 $\pm$ 0.037 \\ 
    NeuralFP & 4            & 1.966 $\pm$ 0.045 \\ 
    \midrule
    Graph Informer  & 1     & 1.827 $\pm$ 0.035 \\ 
    Graph Informer  & 2     & 1.427 $\pm$ 0.024 \\ 
    Graph Informer  & 3     & 1.375 $\pm$  0.016 \\ 
    Graph Informer  & 4     & \textbf{1.348} $\pm$ 0.007 \\ 
    \midrule
    State of the art~\cite{jonas2019rapid}  &    & 1.43 $\pm$ NA \\
    \bottomrule
  \end{tabular}
  \end{sc}
  \end{small}
  \end{center}
\end{table}

In accordance to the most common Lorentzian peak shape assumption in NMR spectroscopy, we minimize the Mean Absolute Error (MAE) \citep{gunther1980nmr}.
For evaluation, we split the data into train-validation-test sets containing 20,000, 2,500, and 3,145 molecules, respectively, and report results over 3 repeats with different splits.
All methods use the same node-level output head, consisting of a $\tanh$ layer followed by a linear output layer.
The MAE loss is minimized only for the atoms that have measurements for the 13C peaks (\emph{i.e.}, the carbon atoms).
For the results, see Table~\ref{tbl:nmr}.

\begin{figure}[t]
  \centering
    \begin{subfigure}[b]{0.25\textwidth}
        \includegraphics[width=\textwidth]{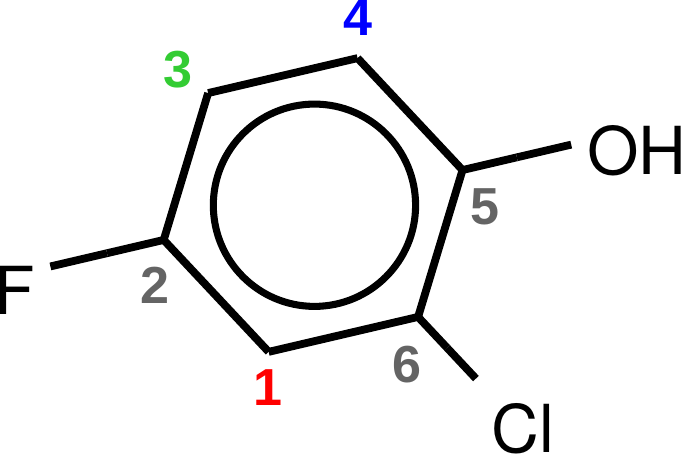}
        \caption{Molecule structure}
        \label{fig:nmr-structure}
    \end{subfigure}
    \hspace{0.5cm}
    \begin{subfigure}[b]{0.45\textwidth}
        \includegraphics[width=\textwidth]{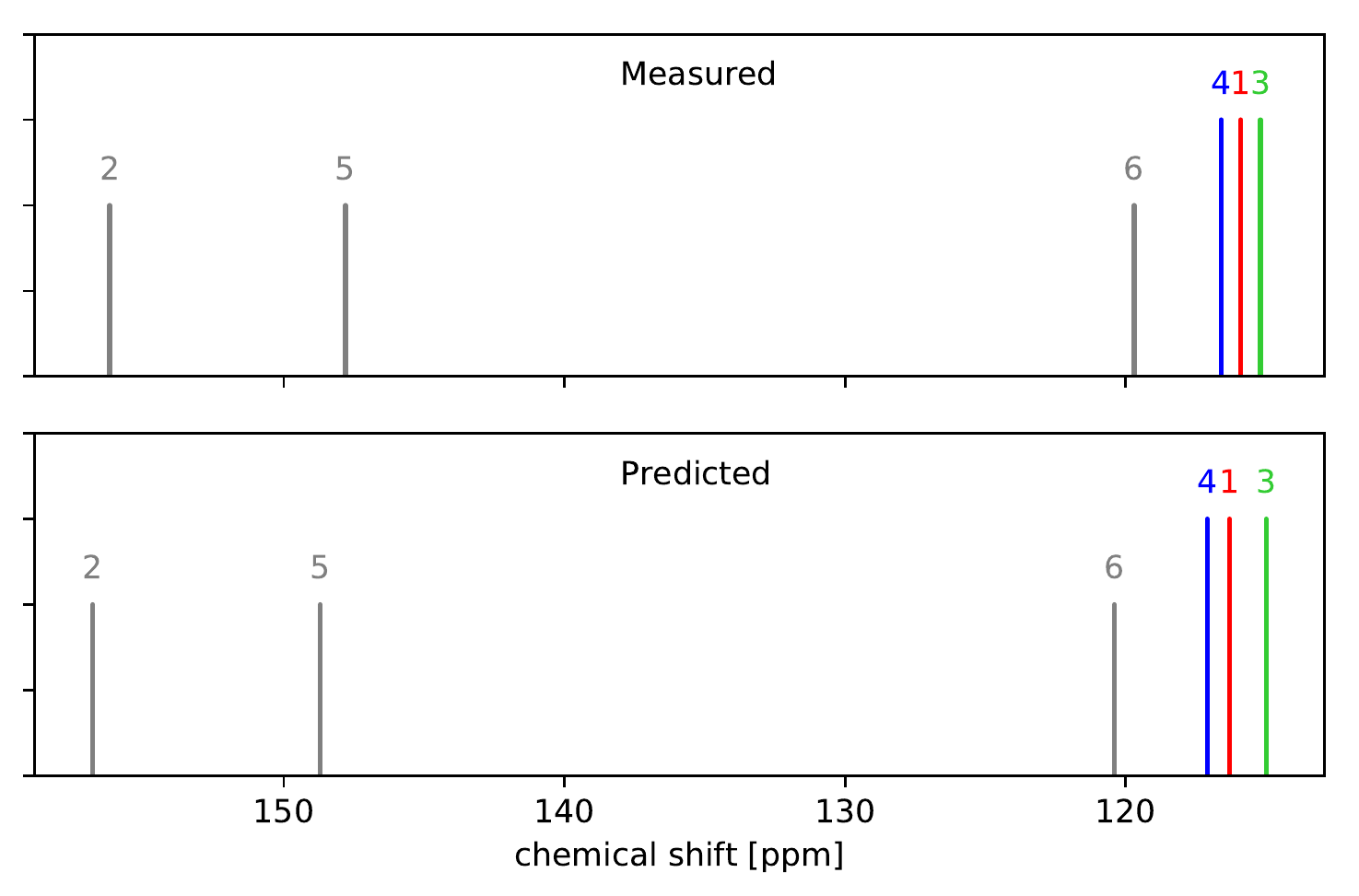}
        \caption{Measured and predicted spectra}
        \label{fig:nmr-predicted}
    \end{subfigure}
  \caption{13C NMR spectrum for 2-chloro-4-fluorophenol. The predicted NMR peaks are in the same order, even in the densely packed region (carbons 4, 1, 3).}
  \label{fig:nmr-example}
\end{figure}

Our proposed \method{} reaches 1.35 MAE, which is also better than the state-of-the-art results from computational chemistry literature, which have reported 1.43 MAE for 13C for the NMRShiftDB2 \cite{jonas2019rapid}.
At these levels of accuracy, many molecules with \emph{densely packed} 13C NMR spectra can be resolved.
This class of molecules with \emph{densely packed} spectra are common for aromatic ring systems, which are prevalent in drug-like compounds.
As an example, see Figure~\ref{fig:nmr-structure} depicting 2-chloro-4-fluorophenol, which has three carbons, labelled 1, 3, 4 with peaks within the 115.2 to 116.6 ppm range.
Figure~\ref{fig:nmr-predicted} displays the predictions from the molecule when it was in the test set. The predictions match closely the true spectrum and, critically, predict the true ordering, which allows us to perfectly match peaks.
For more details and the table with exact prediction values, see Appendix~\ref{app:nmr-error-analysis}.

\begin{figure}[t]
  \centering
    \begin{subfigure}[b]{0.15\textwidth}
        \includegraphics[width=\textwidth]{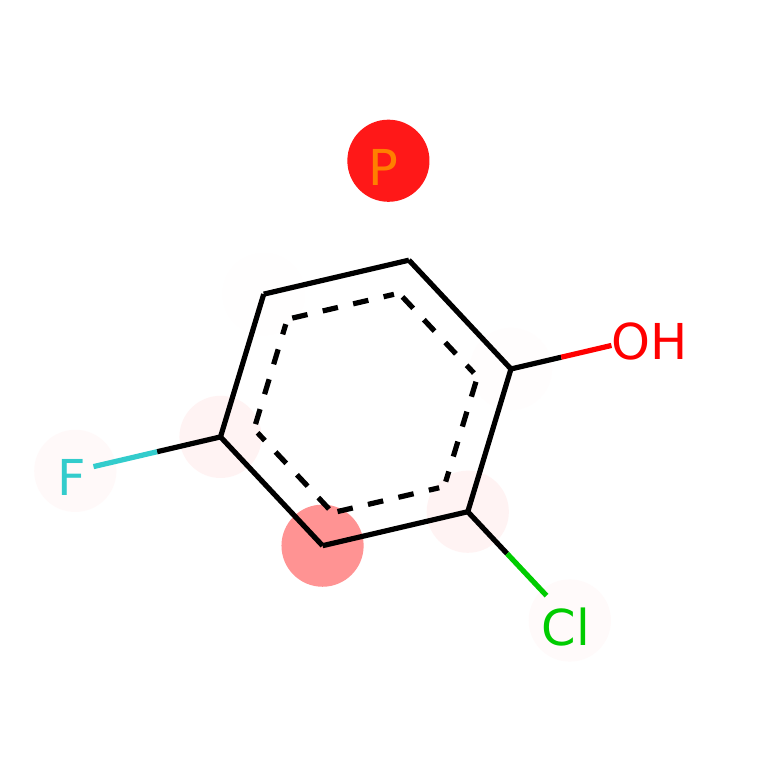}
        \vspace{-0.9cm}
        \caption{layer 1 head 1}
    \end{subfigure}
    \hspace{0.1cm}
    \begin{subfigure}[b]{0.15\textwidth}
        \includegraphics[width=\textwidth]{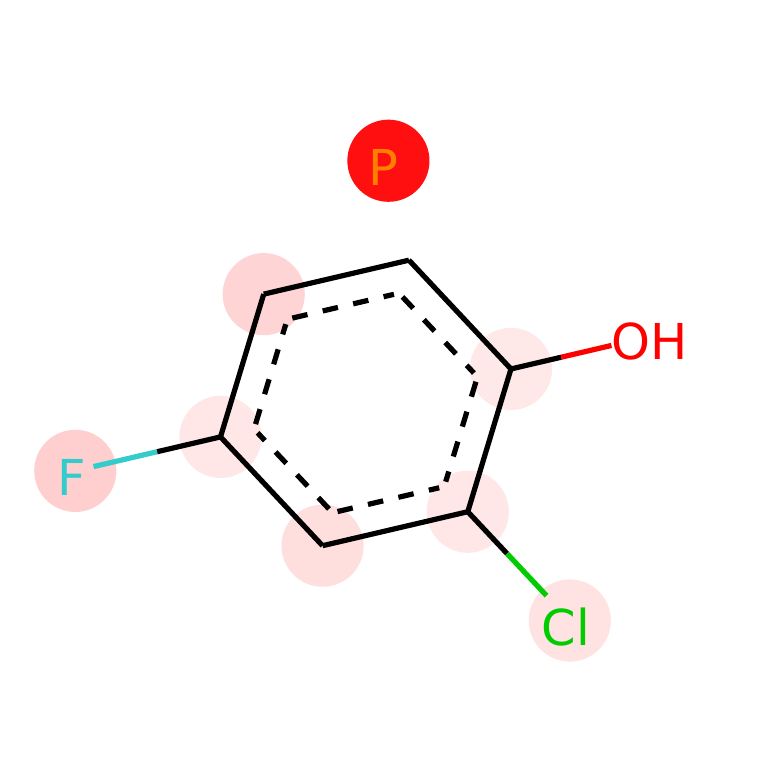}
        \vspace{-0.9cm}
        \caption{layer 1 head 2}
    \end{subfigure}
     \hspace{0.1cm}
    \begin{subfigure}[b]{0.15\textwidth}
        \includegraphics[width=\textwidth]{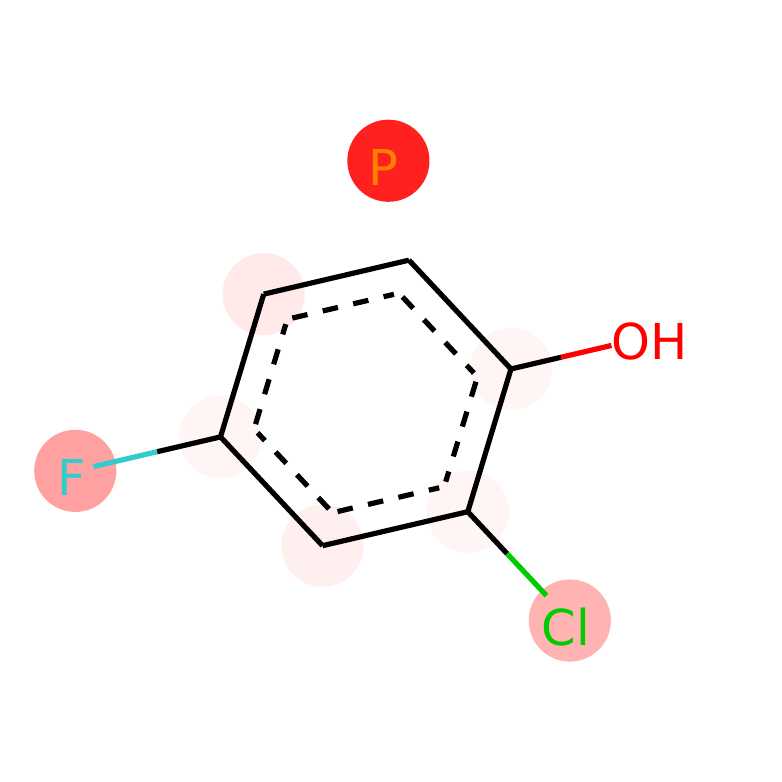}
        \vspace{-0.9cm}
        \caption{layer 1 head 3}
    \end{subfigure}
    
    \begin{subfigure}[b]{0.15\textwidth}
        \includegraphics[width=\textwidth]{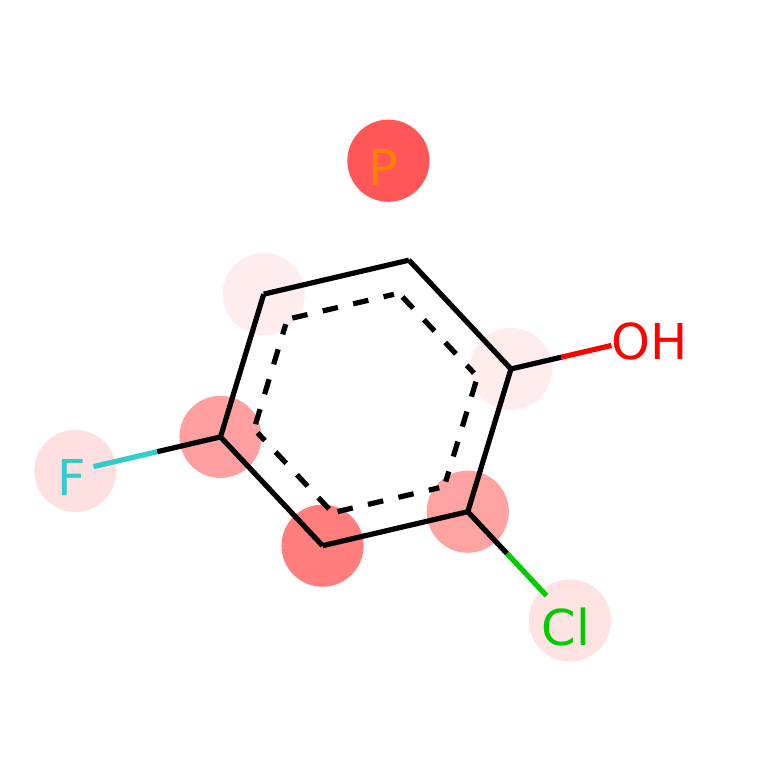}
        \vspace{-0.9cm}
        \caption{layer 2 head 1}
    \end{subfigure}
    \hspace{0.1cm}
    \begin{subfigure}[b]{0.15\textwidth}
        \includegraphics[width=\textwidth]{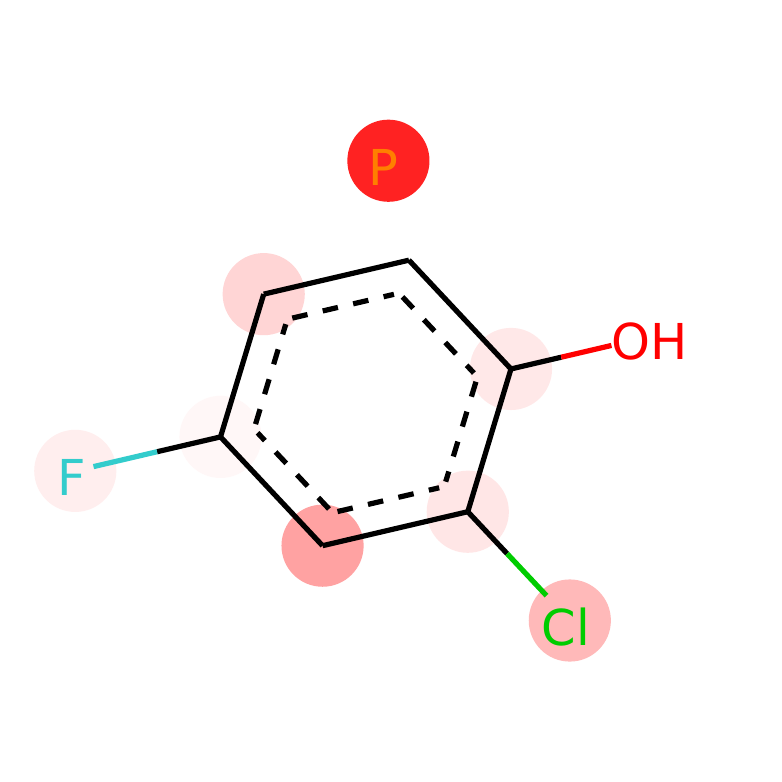}
        \vspace{-0.9cm}
        \caption{layer 2 head 2}
    \end{subfigure}
     \hspace{0.1cm}
    \begin{subfigure}[b]{0.15\textwidth}
        \includegraphics[width=\textwidth]{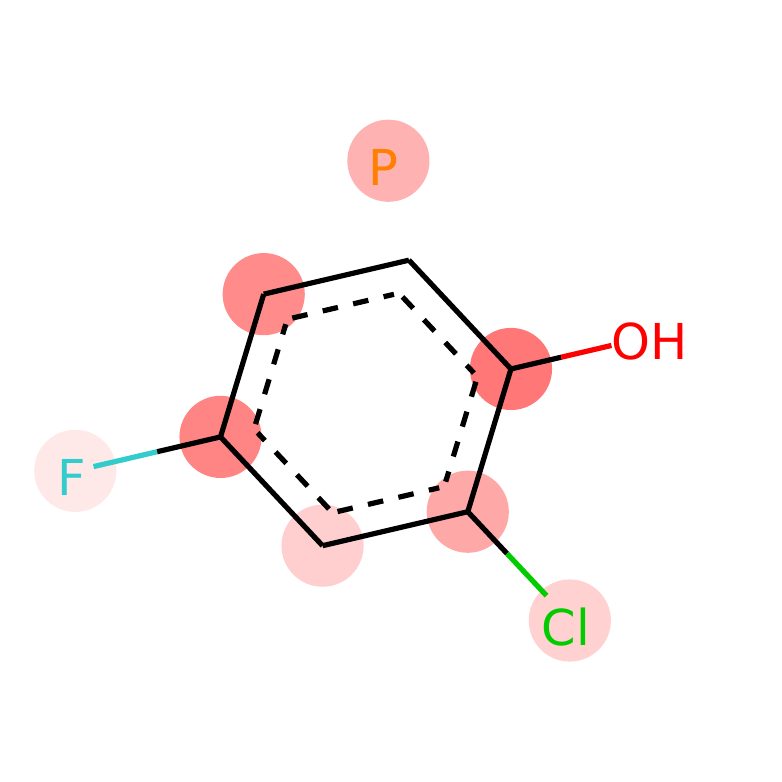}
        \vspace{-0.9cm}
        \caption{layer 2 head 3}
    \end{subfigure}
    \caption{Attention probabilities from carbon 1 (bottom in the figures) in the 2-chloro-4-fluorophenol molecule. We show selected three heads from layer 1 and layer 2. For the full list see Appendix~\ref{app:attention}.}
    \label{fig:attn-prob}
\end{figure}

For a \method{} with 3 layers and 6 heads all having radius 2, we depict the attention probabilities in Figure~\ref{fig:attn-prob} for the bottom carbon (number 1 in Figure~\ref{fig:nmr-structure}) of same 2-chloro-4-fluorophenol compound.
We can see that all heads in Layer 1 strongly use the pool node as a reference.
Also, we can see that in Layer 1 the first head focuses on the atom itself, the second head attends to all neighbors within radius 2, and the third head focuses on the halogen atoms.

\subsection{Results for graph-level tasks}
\label{sec:chembl}
We benchmark the methods on two drug bioactivity data sets: ChEMBL v23 by \citet{bento2014chembl} (94 tasks, 87,415 molecules, and 349,192 activity points) and Tox21 from MoleculeNet by \citet{wu2018moleculenet} (12 tasks, 7,831 molecules, and 77,946 activity points).
For ChEMBL, we used a repeated holdout-validation, and for Tox21 we used the single train-validation-test split given by MoleculeNet (see Appendix~\ref{app:tox21} for the details).
In the case of ChEMBL, because of well-known series effect in drug design, allocating molecules randomly to the training and test sets results in overly optimistic and inconsistent performance estimates \citep{simm2018repurposing} because highly similar molecules end up in both training and test sets. Therefore, we apply \emph{clustered} hold-out validation (See Appendix~\ref{app:clustered}).  
We allocate the clusters (rather than individual molecules) to the training (60\%), validation (20\%), and test sets (20\%). We repeat this split 5 times.

\begin{table}[t]
\caption{AUC-ROC scores for multi-task datasets ChEMBL and Tox21 of MoleculeNet. The AUCs are calculated for each task and averaged.}
\label{tbl:chembl-tox21}
\vskip 0.15in
\begin{center}
\begin{small}
\begin{sc}
\begin{tabular}{lcc}
\toprule
Method    & ChEMBL & Tox21 \\
\midrule
NeuralFP              & $0.826 \pm 0.005$    & 0.829  \\
GCN                   & $0.814 \pm 0.005$    & 0.809  \\
GIN                   & $0.816 \pm 0.004$    & 0.810  \\
GGNN                  & $0.756 \pm 0.014$    & 0.809  \\
NeuralMP              & $0.707 \pm 0.005$    & 0.804  \\
Weave                 & $0.830 \pm 0.006$    & 0.822  \\
\midrule
Graph Informer  & $\textbf{0.839} \pm 0.003$ & \textbf{0.848} \\
\bottomrule
\end{tabular}
\end{sc}
\end{small}
\end{center}
\vskip -0.1in
\end{table}

Our and all adapted methods (GCN, GIN and GGNN) use a graph pooling layer consisting of a linear layer, ReLU, mean pooling, and a linear layer.
We minimize the cross-entropy loss and Table~\ref{tbl:chembl-tox21} reports the AUC-ROC averages over tasks for each data set.
\method{} outperforms all baseline methods in both ChEMBL and Tox21 data sets.

\section{Conclusion}
In this work, we proposed a route-based multi-head attention mechanism that allows the attention to use relative position and the type of connection between the pair of nodes in the graph.
Because of its route features, our approach can incorporate information from nondirect neighbors efficiently---similarly to the use of dilated filters in CNNs.
In our theoretical analysis, we showed that a variant of \method{} is provably as powerful as the WL $\mathrm{dim}=1$ test.
Our empirical evaluation demonstrated that the proposed method is suitable both for node-level and graph-level prediction tasks, and delivers significant improvements over existing approaches in 13C NMR spectrum and drug bioactivity prediction.

\appendices

\section{Node and route features}
In addition to the following features, we fed the one-hot encoding of the atom types by concatenating it to the node feature vector, see Table~\ref{tbl:node_features}.
The atom types that occurred in the two data sets are \{C, N, O, Cl, F, S, I, Br, P, B, Zn, Si, Li, Na, Mg, K\}.
As the route features (Table~\ref{tbl:route_features}) contain all edge labels (single, double, triple and aromatic bond type), all information about the graph topology is retained.
\label{app:features}
\begin{table}[h!]
  \caption{Node features}
  \label{tbl:node_features}
  \centering
  \begin{tabular}{cl}
    \toprule
    Position     & Description \\
    \midrule
    0-2 & Formal charge, one-hot encoded \{-1, 0, +1\} \\
    3-7 & Hybridization state, one-hot encoded \{s, sp, sp2, sp3\} \\
    8-13 & Explicit valence, one-hot encoded integer, between 0 and 5 \\
    14 & Aromaticity, binary \\
    15 & Whether it is in a ring size 3, binary \\
    16 & Whether it is in a ring size 4, binary \\
    17 & Whether it is in a ring size 5, binary \\
    18 & Whether it is in a ring size 6, binary \\
    19 & Whether it is in any ring, binary \\
    20 & Partial charge, computed by Gasteiger method, real number \\
    21 & Whether it is a H-acceptor, binary \\
    22 & Whether it is a H-donor, binary \\
    23 & Whether it is an R stereo center, binary \\
    22 & Whether it is an S stereo center, binary \\
    \bottomrule
  \end{tabular}
\end{table}

\begin{table*}[t]
  \caption{Route features}
  \label{tbl:route_features}
  \centering
  \begin{tabular}{cl}
    \toprule
    Position     & Description \\
    \midrule
    0-8 & Bond distance, binned, [0, 1, 2, 3, 4, 5 $\leq$ d $\leq$ 6, 7 $\leq$ d $\leq$ 8, 9 $\leq$ d $\leq$ 12, 13 $\leq$ d  ] \\
    9 & Whether the shortest pure conjugated path containing at most 4 bonds, binary \\
    10 & Whether the shortest pure conjugated path containing at least 5 bonds, binary \\
    11 & Whether there is a route containing only up to 13 single bonds, binary \\
    12 & Whether there is a route containing only up to 13 double bonds, binary \\
    13 & Triple bond, binary \\
    14 & Whether there is a shortest path which is a pure conjugated path, binary \\
    15 & Whether the endpoints are in the same ring, binary \\
    16 & Single bond, binary \\
    17 & Double bond, binary \\
    18 & Aromatic bond, binary \\
    \bottomrule
  \end{tabular}
\end{table*}

\section{Baselines used in experiments}
\label{app:baselines}

    \begin{center}
    \begin{footnotesize}
    \begin{tabular}{r|c|l}
    \toprule
    Method & Paper & Source code  \\
    \midrule
    GGNN & \cite{li2015gated} & \url{https://github.com/JamesChuanggg/ggnn.pytorch/} \\
    NeuralFP & \cite{duvenaud2015convolutional} & Our implementation based on the paper \\
    GCN & \cite{kipf2016semi} & \url{https://github.com/tkipf/pygcn.git}  \\
    Weave & \cite{kearnes2016molecular} & \url{https://github.com/deepchem/deepchem/} \\
    GAT & \cite{velivckovic2017graph} & \url{https://github.com/PetarV-/GAT} \\
    NeuralMP & \cite{gilmer2017neural} & \url{https://github.com/brain-research/mpnn} \\
    GIN & \cite{xu2018powerful} & Adapted GCN code to use sum instead of \\
     & & average and added epsilon \\
    \bottomrule
    \end{tabular}
    \end{footnotesize}
    \end{center}

\section{Preprocessing of the ChEMBL data set}
\label{app:preprocessing_chembl}

\subsection{Extracting human drug--protein interaction data}
We extracted IC50 values for \emph{Homo sapiens} protein targets from ChEMBL version 23 SQLite release~\citep{bento2014chembl}  using the following SQL query, and computed the negative log10 of IC50 in nM, which we refer to as pIC50. Extreme values that cannot correspond to meaningful real measurement results are treated as database errors and filtered out (IC50 $< 10^{-5}$ or $10^9 \leq$ IC50). The ChEMBL SQL query is shown in Table~\ref{tbl:chembl-sql-query}.

\begin{table*}[t]
  \caption{SQL query for extracting bioactivity data for chemical compounds on human proteins.}
  \label{tbl:chembl-sql-query}
  \centering
\begin{tabular}{ c }
\begin{minipage}{0.8\textwidth}
\begin{Verbatim}[fontsize=\small]

SELECT molecule_dictionary.chembl_id as cmpd_id,
    target_dictionary.chembl_id as target_id, 
CASE activities.standard_units
    WHEN 'nM' THEN activities.standard_value
    WHEN 'ug.mL-1' THEN
        activities.standard_value / compound_properties.full_mwt * 1E6
    END ic50,
CASE activities.standard_relation 
    WHEN '<'  THEN '<'
    WHEN '<=' THEN '<'
    WHEN '='  THEN '='
    WHEN '>'  THEN '>'
    WHEN '>='  THEN '>' 
    ELSE 'drop' END relation
  FROM molecule_dictionary 
  JOIN activities on activities.molregno == molecule_dictionary.molregno 
  JOIN assays on assays.assay_id == activities.assay_id 
  JOIN target_dictionary on target_dictionary.tid == assays.tid
  JOIN compound_properties on
    compound_properties.molregno = molecule_dictionary.molregno
  WHERE target_dictionary.organism='Homo sapiens' AND
    target_dictionary.target_type='SINGLE PROTEIN' AND
    activities.standard_type = 'IC50' AND
    activities.standard_units IN  ('nM','ug.mL-1') and
    relation != 'drop' and
    ic50 < 10e9 AND
    ic50 >= 10e-5
\end{Verbatim}
\end{minipage}
\end{tabular}
\end{table*} 


\subsection{Filtering}
First, we sanitized the molecules by removing inactive ions (Cl-, Na+, K+, Br-, I-) from the structure.
Then, we removed all non-drug-like molecules containing heavy metals or more than 80 non-hydrogen atoms. We thresholded the activity for each protein with four pIC50 thresholds (5.5, 6.5, 7.5, and 8.5) taking into account the \emph{relation} column.
We then kept protein--threshold pairs that had measurements at least on 2,500 compounds containing at least 200 positive and 200 negative samples.
This filtering process resulted in 94 protein--threshold pairs as binary classification tasks on 87,415 molecules. The matrix contains 349,192 activity points.

\subsection{Clustered hold-out validation}
\label{app:clustered}
We clustered the molecules with similarity threshold 0.6 Tanimoto on ECFP6 features computed using RDKit \citep{rdkit}. For clustering, we used the leader-follower method \citep{hartigan1975clustering}. We then assigned each cluster to the training, test, or validation sets randomly. We adapted the clustered validation pipeline from \citet{simm2018repurposing}.

\section{Tox21 experimental setup}
\label{app:tox21}
We used the Tox21 split by MoleculeNet, with a total of 7,831 molecules divided into 6,264 train, 783 validation, and 784 test compounds.
Each method and each hyperparameter was trained 3 times and the best performing configuration was chosen using its validation AUC-ROC. All runs also used early stopping with validation AUC-ROC.

\section{NMR data set}
\label{app:nmr-dataset}
To create the data set, we used 13C NMR spectra from NMRShiftDB2, available from \url{https://nmrshiftdb.nmr.uni-koeln.de/}.
We kept only molecules where all carbon atoms where assigned a 13C peak position and removed any compounds that failed RDKit~\citep{rdkit} sanitization or Gasteiger partial charge calculation.
This resulted in 25,645 molecules.
The peak positions are measured in parts per million (ppm) and ranges from approximately -10 to 250 ppm. Note that the ppm value measures the shift of the peak relative to a specific reference carbon atom.
In our data set, the mean and standard deviation are 95 and 52 ppm.

\section{Description of the NMR application}
\label{app:nmr-error-analysis}

In molecules, the chemical environment of the atomic nuclei has an influence on their measured resonance frequency, the NMR peak position.
This influence is called the chemical shift, and measured in parts per million (ppm).
The nominal value of this shift is independent from the measurement equipment and can be used to compare chemical environments between molecules measured in different laboratories.

We are specifically interested in the following scenario. An organic chemist intends to synthesize a compound. After the synthesis and purification steps, the identity of the product needs to be validated.
The chemist can measure the 13C NMR spectrum of the product, and compare it to the predicted spectrum from the structure of this compound.
To compare the two spectra, the correct peaks corresponding to the same atom need to be aligned.
There is no way to tell from a 13C NMR spectrum alone which peak corresponds to which atom. Therefore, wrongly ordered peaks in the prediction would result in an erroneous assignment.

As in the example in the main text, we use the 13C NMR spectrum of the compound 2-chloro-4-fluorophenol to illustrate the assignment problem.
The measured and predicted chemical shifts are reported in Table~\ref{tbl:spectrum}.
As can be seen in Figure~\ref{fig:nmr-example}, the three unsubstituted aromatic carbons (1,3,4) have very close peaks in the 115.2 to 116.6 ppm region. 
For the sake of completeness note that the intensity of the peaks (value of the $y$-axis) in the plot are directly related to the number of direct hydrogen neighbors of the carbon, and therefore does not need to be predicted \citep{doddrell1982distortionless}. 

\begin{table}[t]
  \caption{13C NMR spectrum assignment of 2-chloro-4-fluorophenol.}
  \label{tbl:spectrum}
  \centering
  \begin{tabular}{rrrrrrrrrrr}
    \toprule
    Carbon      & 2 & 5 & 6 & 4 & 1 & 3  \\
    \midrule
    Measured   & 156.2 & 147.8 & 119.7 & 116.6 & 115.9 & 115.2  \\
    Predicted  & 156.8 & 148.7 & 120.4 & 117.1 & 116.3 & 115.0  \\
    \midrule
    Abs. Error & 0.6 & 0.9 & 0.7 & 0.5 & 0.4 & 0.2 \\
    \bottomrule
  \end{tabular}
\end{table}

\section{Improving the convergence of \method{}}
\label{app:gradient-issue}
Precisely following the strategy used for transformer networks~\citep{vaswani2017AttentionAllYou}, we get the following architecture for graphs: 
\begin{align}
    T  &= \mathrm{LayerNorm}(H + \mathrm{Linear}(\mathrm{RouteMHSA}(H)))
    \label{eq:app:layernorm-route}
    \\
    H' &= \mathrm{LayerNorm}(T + \mathrm{FFN}(T)),
    \\
    \mathrm{FFN}(x) &= W_2 \mathrm{ReLU}(W_1 x + b_1) + b_2,
\end{align}
and $H'$ is the output of the block (\emph{i.e.,} updated hidden vectors).

In our experiments, we observed that the output of $\mathrm{RouteMHSA}(H)$ dominates the input $H$ in Eq.~\ref{eq:app:layernorm-route}.
This causes convergence issues as normalizing this sum by the $\mathrm{LayerNorm}$ causes the gradient through the input $H$ branch to vanish.
This was the main motivation for introducing the pure-residual style network described in Section~\ref{sec:architectures}.

\section{Isomorphism test}

Randomly initialized untrained 1-layer injective \method{} was used to separate regular graphs when using histogram of route counts as route features.
For the histogram, we used routes up to length ($k=4$) (\emph{i.e.}, the feature vector for each pair of nodes contains the counts of routes of length 1, 2, 3, and 4).
After the embeddings for all nodes have been computed the embedding for the graph is their sum. 
Finally the two graphs were considered different if their embeddings differed more than $10^{-4}$.
We used the following setup:

\begin{table}[h!]
\begin{center}
\begin{sc}
\begin{tabular}{l|c}
\toprule
Parameter & Value  \\
\midrule
Hidden size         & 8 \\
Number of heads     & 4 \\
Head radius         & 8 (can see the whole graph) \\
Route feature size  & 4 \\
Key size            & 2 \\
Route key size      & 2 \\
Input of size       & 1 (constant 1.0 for all nodes).\\
\bottomrule
\end{tabular}
\end{sc}
\end{center}
\end{table}

\subsection{Regular 8-node 3-degree graphs}
\label{app:beyond-wl}
\begin{figure}[t]
  \centering
    \begin{subfigure}[b]{0.2\textwidth}
        \includegraphics[width=\textwidth]{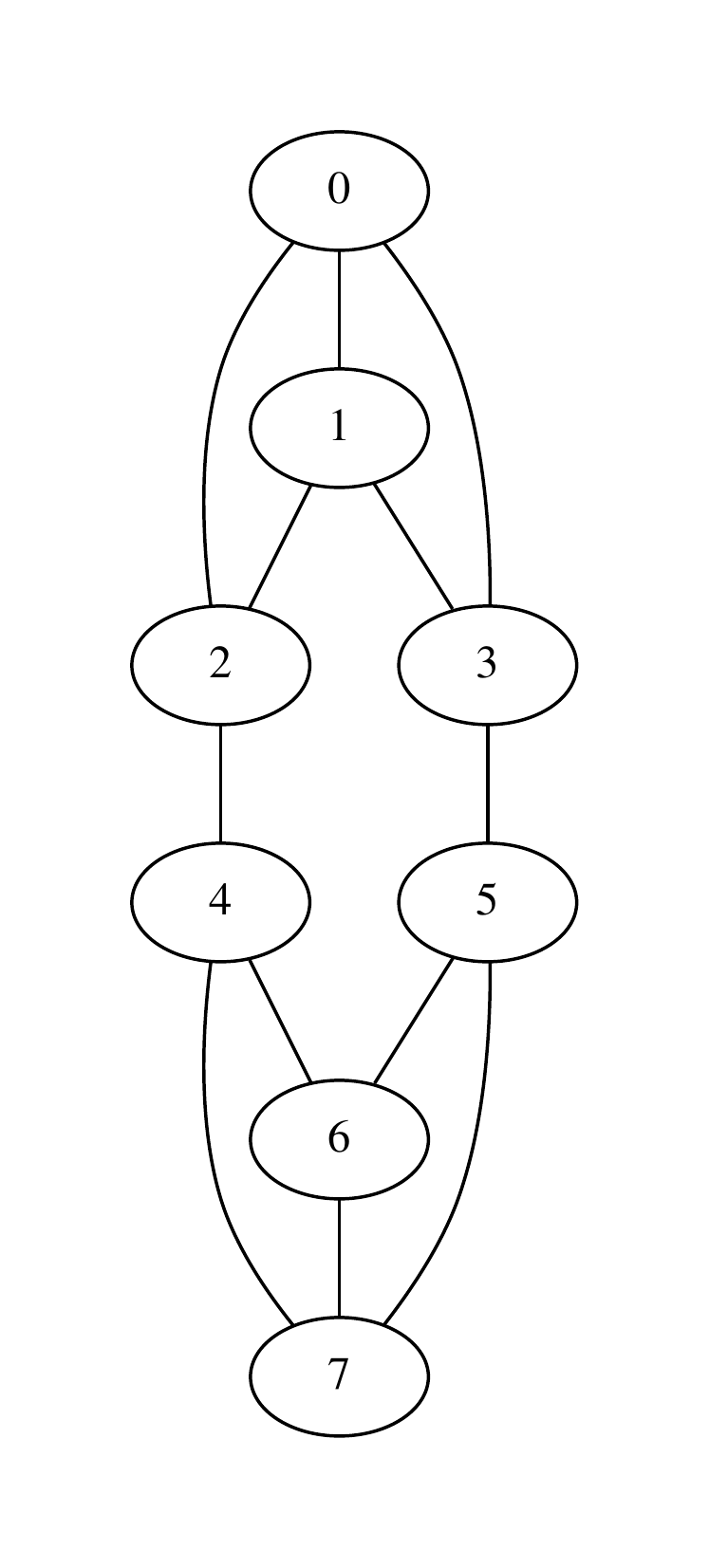}
        \caption{G1}
        \label{fig:G1}
    \end{subfigure}
    \begin{subfigure}[b]{0.2\textwidth}
        \includegraphics[width=\textwidth]{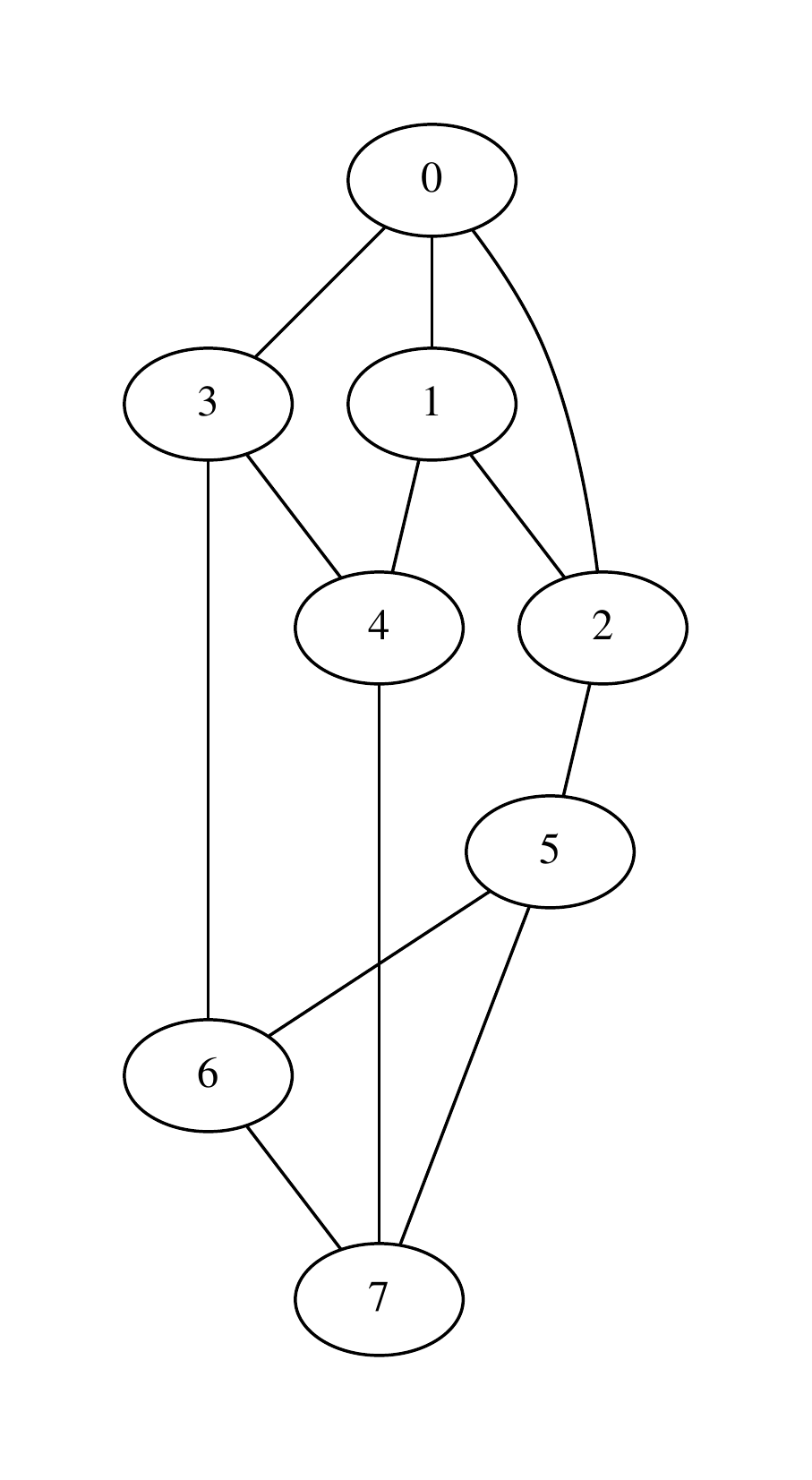}
        \caption{G2}
        \label{fig:G2}
    \end{subfigure}
    \hspace{0.5cm}
    \begin{subfigure}[b]{0.2\textwidth}
        \includegraphics[width=\textwidth]{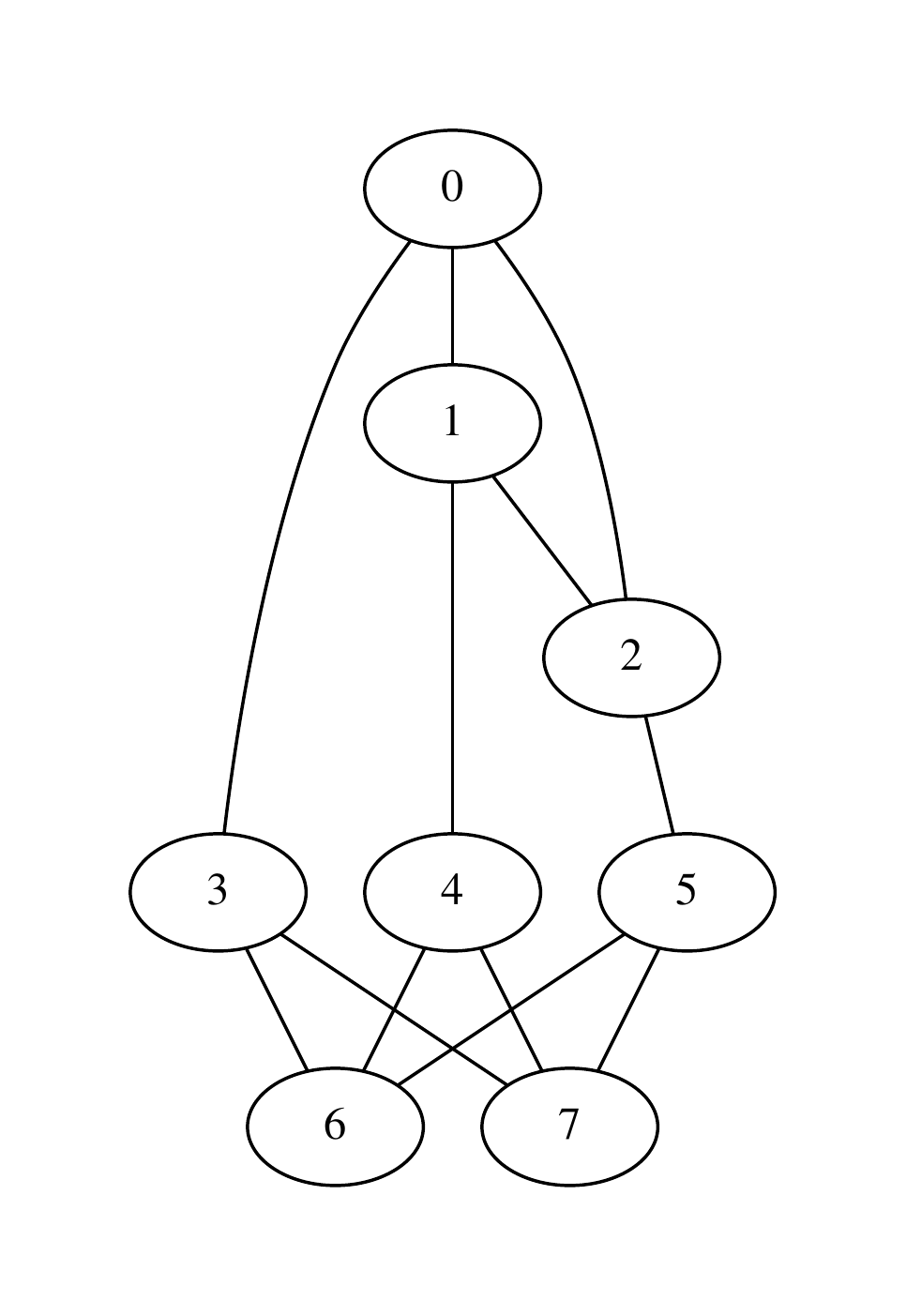}
        \caption{G3}
        \label{fig:G3}
    \end{subfigure}
    \begin{subfigure}[b]{0.2\textwidth}
        \includegraphics[width=\textwidth]{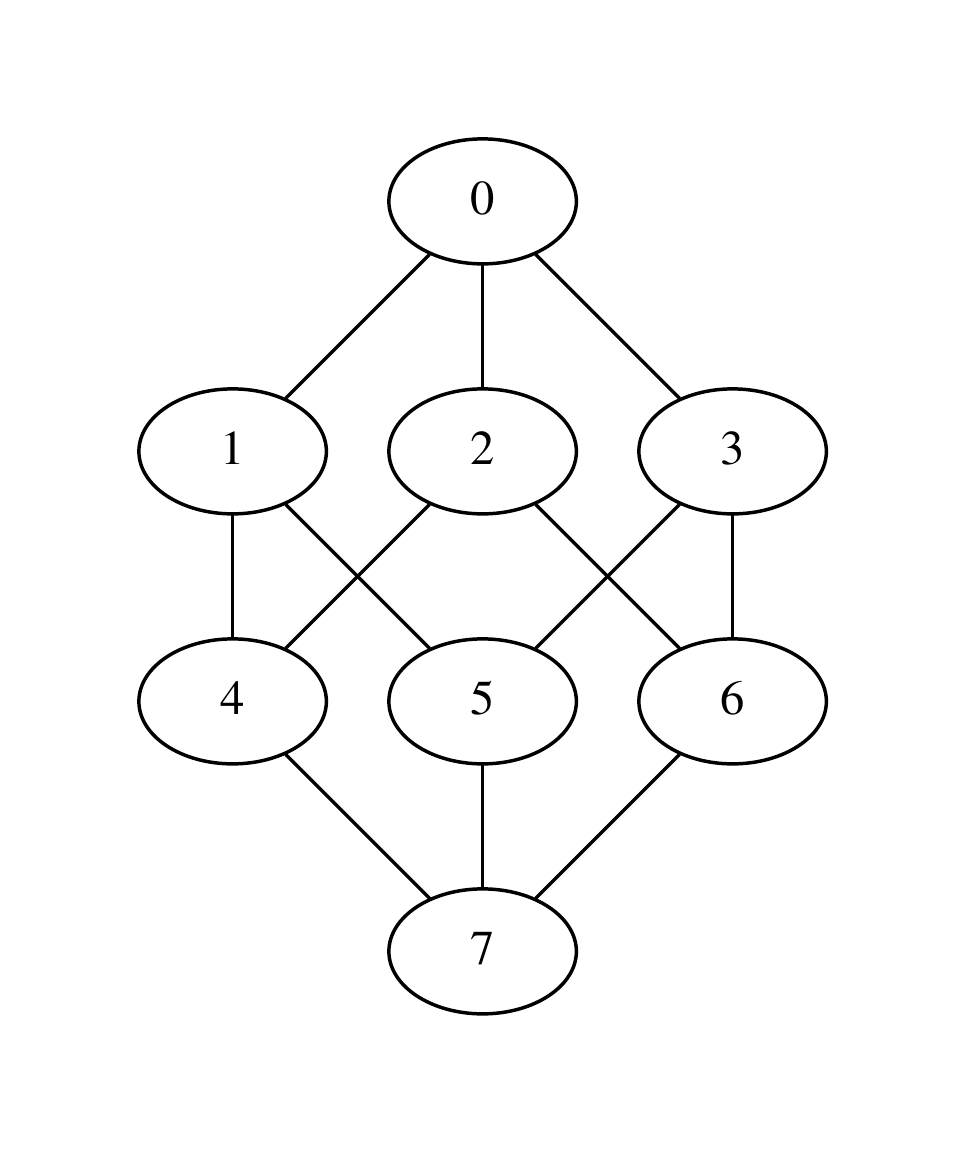}
        \caption{G4 (cubane)}
        \label{fig:G4}
    \end{subfigure}
    \begin{subfigure}[b]{0.2\textwidth}
        \includegraphics[width=\textwidth]{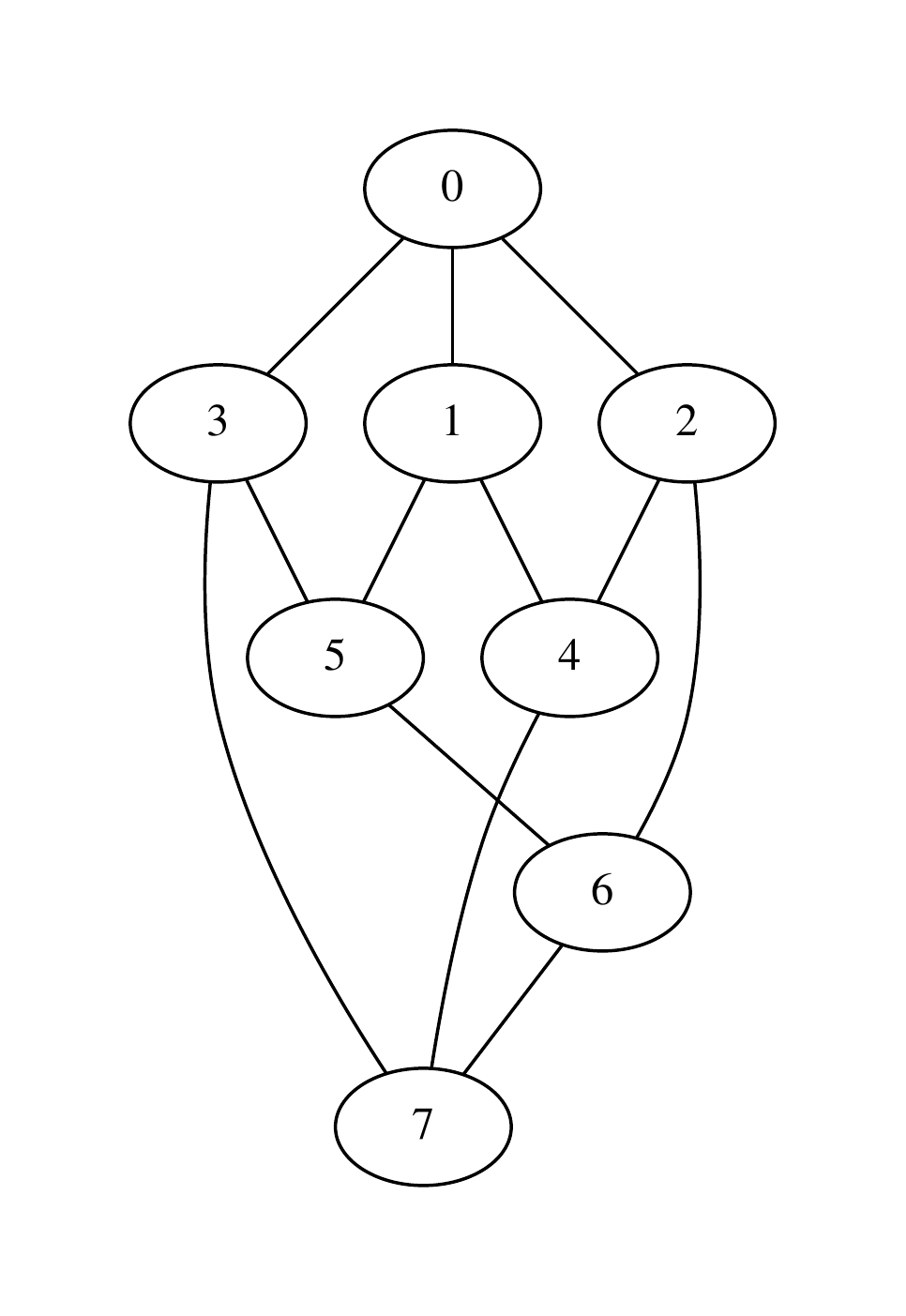}
        \caption{G5}
        \label{fig:G5}
    \end{subfigure}
  \caption{All 5 regular 8-node 3-degree graphs.}
  \label{fig:RegN83D}
\end{figure}

All 5 regular 8-node 3-degree graphs are depicted in Figure~\ref{fig:RegN83D}.
These graphs are not distinguishable by the Weisfeiler-Lehman test with $\mathrm{dim}=1$ and consequently by many graph neural networks.

\vfill
\newpage

\section{Attention plots}
\label{app:attention}
\begin{figure}[h!]
  \centering
    \begin{subfigure}[b]{0.15\textwidth}
        \includegraphics[width=\textwidth]{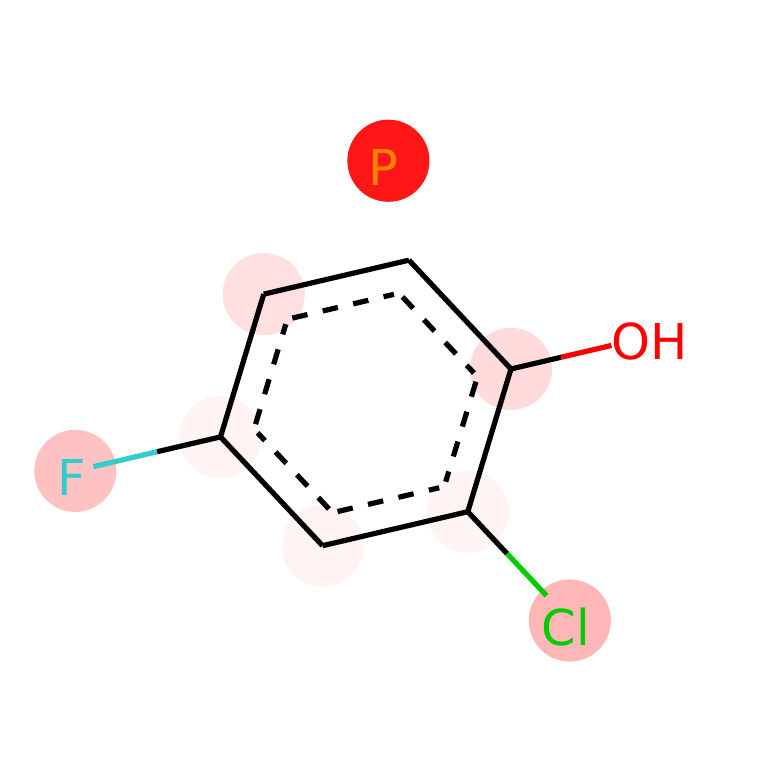}
    \end{subfigure}
    \hspace{0.1cm}
    \begin{subfigure}[b]{0.15\textwidth}
        \includegraphics[width=\textwidth]{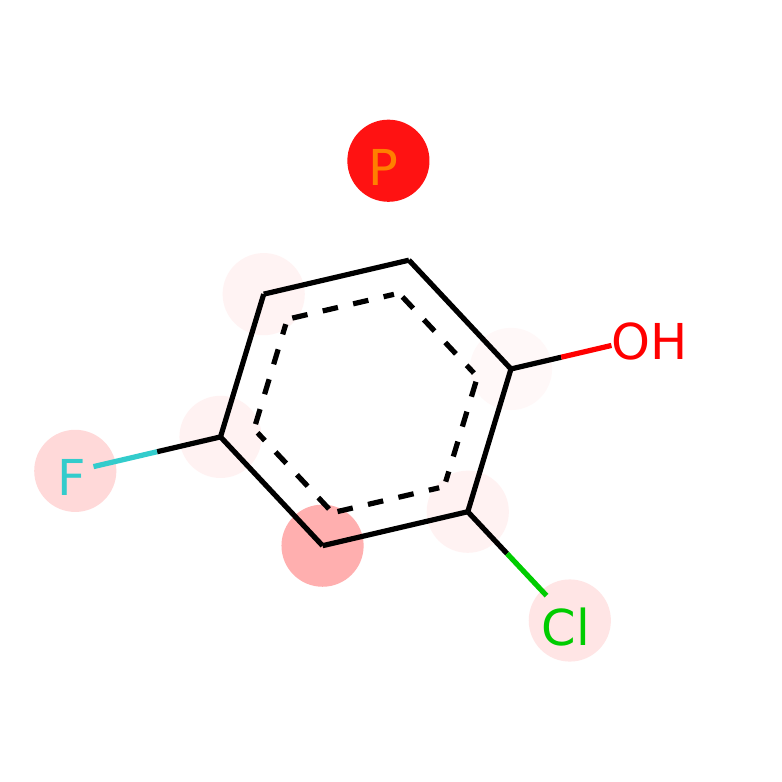}
    \end{subfigure}
    \hspace{0.1cm}
    \begin{subfigure}[b]{0.15\textwidth}
        \includegraphics[width=\textwidth]{figures/test_head_layer_0_head_2.pdf}
    \end{subfigure}
    \hspace{0.1cm}
    \begin{subfigure}[b]{0.15\textwidth}
        \includegraphics[width=\textwidth]{figures/test_head_layer_0_head_3.pdf}
    \end{subfigure}
    \hspace{0.1cm}
    \begin{subfigure}[b]{0.15\textwidth}
        \includegraphics[width=\textwidth]{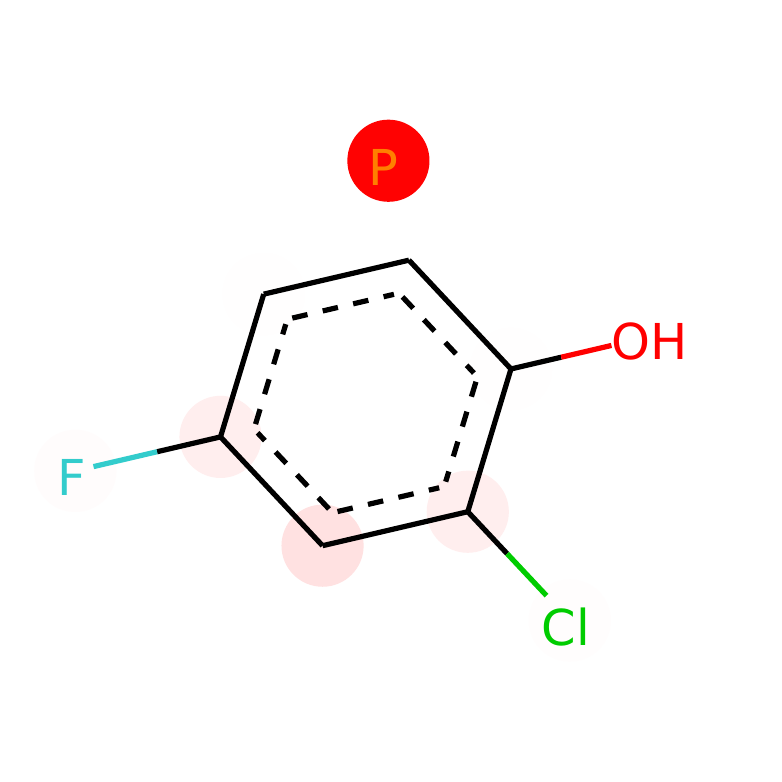}
        
    \end{subfigure}
    \hspace{0.1cm}
    \begin{subfigure}[b]{0.15\textwidth}
        \includegraphics[width=\textwidth]{figures/test_head_layer_0_head_5.pdf}
        
    \end{subfigure}
    \begin{subfigure}[b]{0.15\textwidth}
        \includegraphics[width=\textwidth]{figures/test_head_layer_1_head_0.pdf}
        
    \end{subfigure}
    \hspace{0.1cm}
    \begin{subfigure}[b]{0.15\textwidth}
        \includegraphics[width=\textwidth]{figures/test_head_layer_1_head_1.pdf}
        
    \end{subfigure}
    \hspace{0.1cm}
    \begin{subfigure}[b]{0.15\textwidth}
        \includegraphics[width=\textwidth]{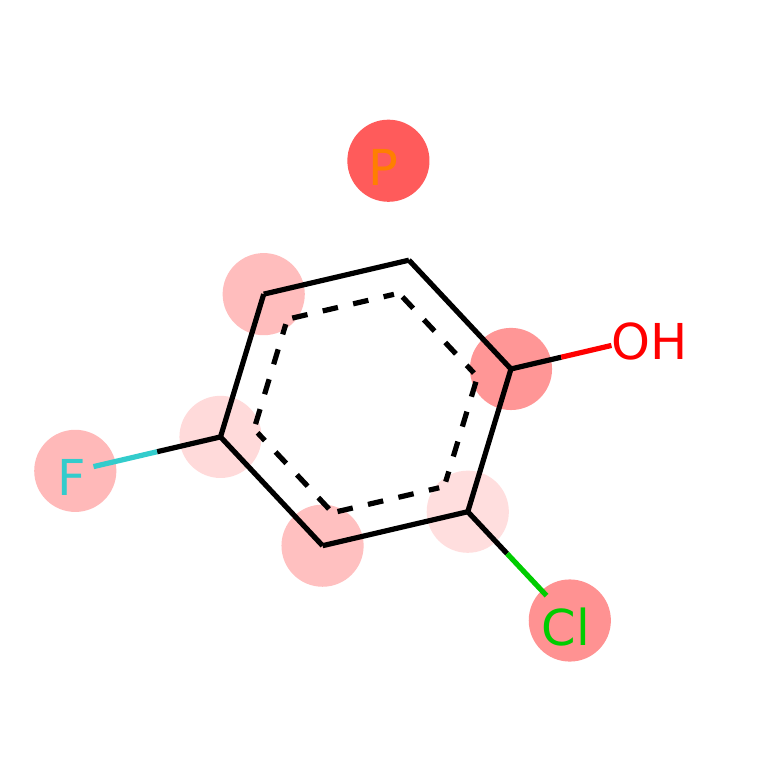}
        
    \end{subfigure}
    \hspace{0.1cm}
    \begin{subfigure}[b]{0.15\textwidth}
        \includegraphics[width=\textwidth]{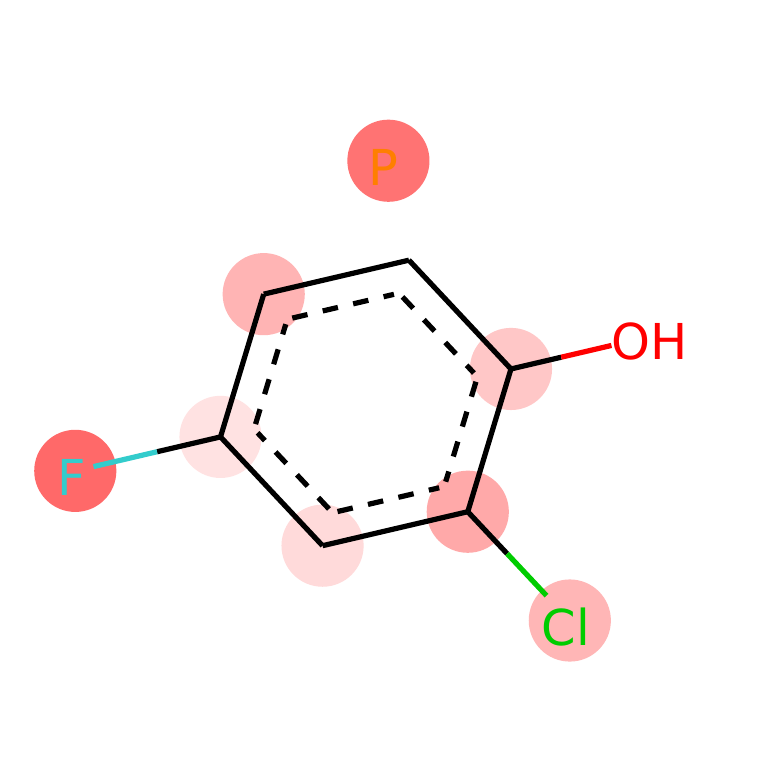}
        
    \end{subfigure}
    \hspace{0.1cm}
    \begin{subfigure}[b]{0.15\textwidth}
        \includegraphics[width=\textwidth]{figures/test_head_layer_1_head_4.pdf}
        
    \end{subfigure}
    \hspace{0.1cm}
    \begin{subfigure}[b]{0.15\textwidth}
        \includegraphics[width=\textwidth]{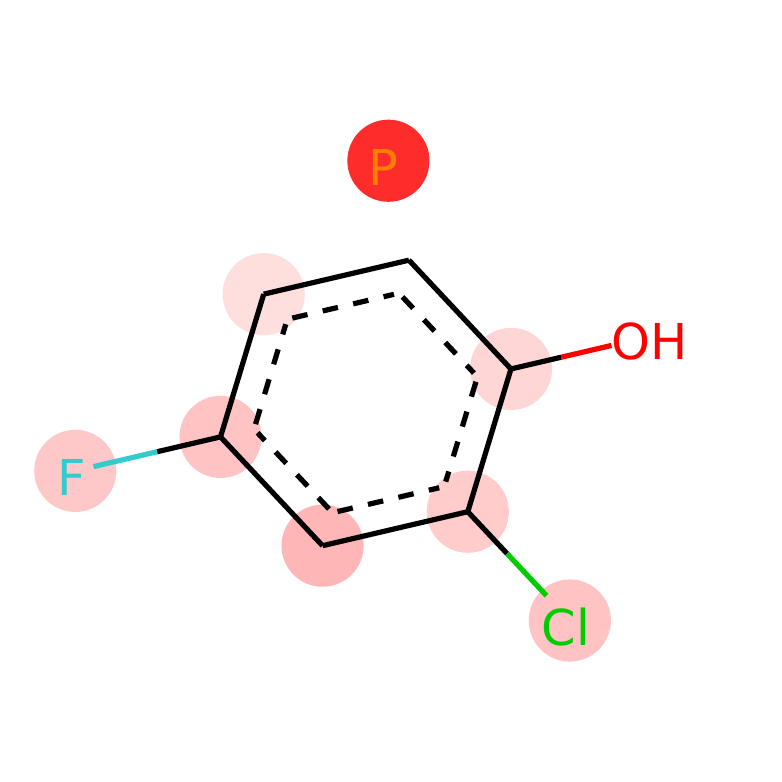}
        
    \end{subfigure}
    \begin{subfigure}[b]{0.15\textwidth}
        \includegraphics[width=\textwidth]{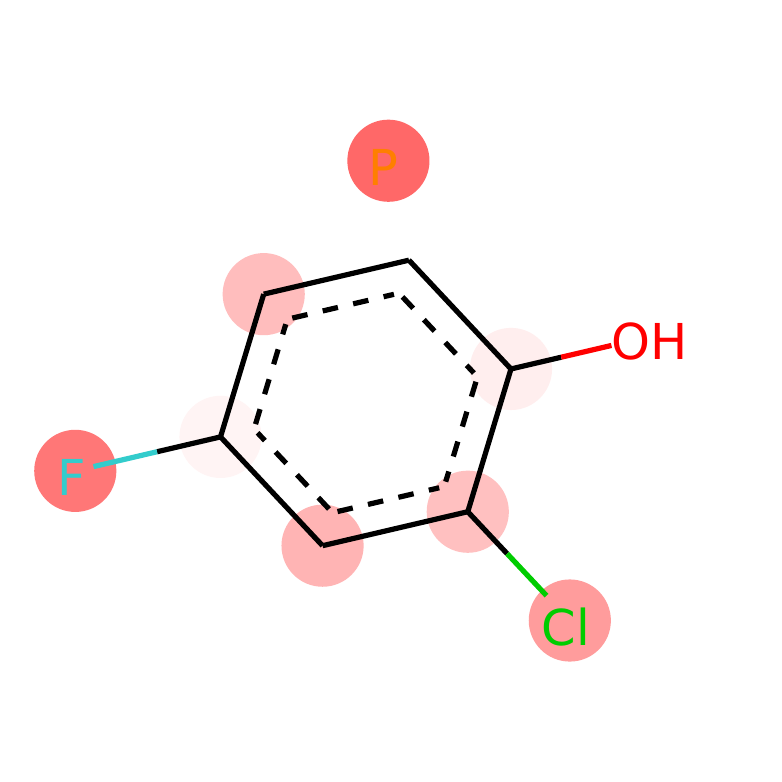}
        
    \end{subfigure}
    \hspace{0.1cm}
    \begin{subfigure}[b]{0.15\textwidth}
        \includegraphics[width=\textwidth]{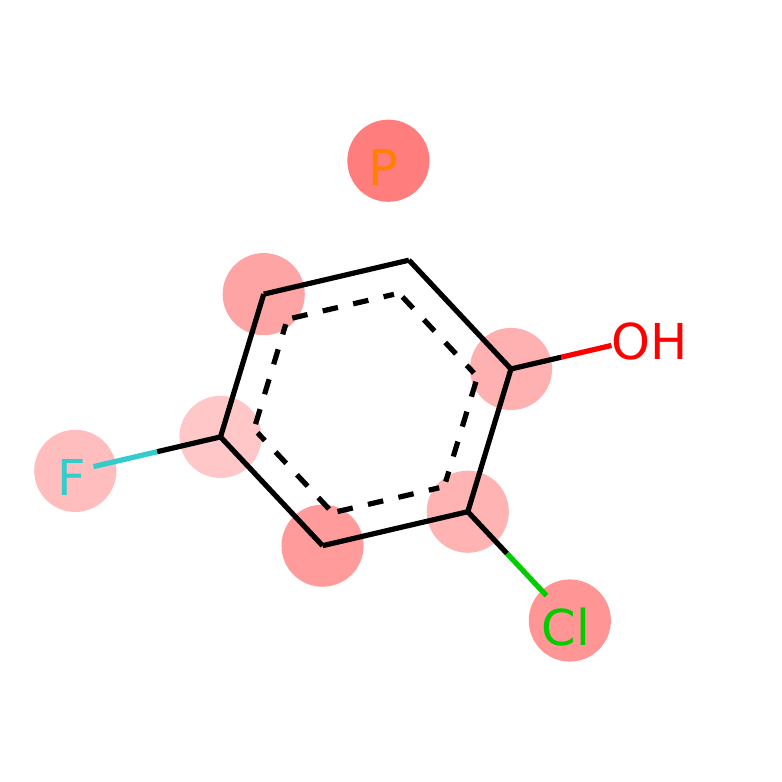}
        
    \end{subfigure}
    \hspace{0.1cm}
    \begin{subfigure}[b]{0.15\textwidth}
        \includegraphics[width=\textwidth]{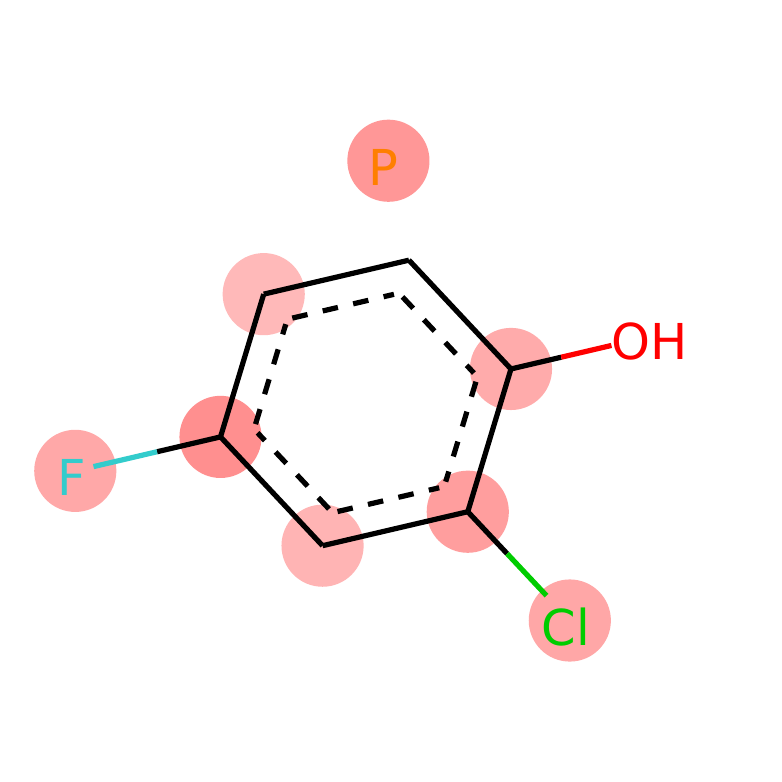}
        
    \end{subfigure}
    \hspace{0.1cm}
    \begin{subfigure}[b]{0.15\textwidth}
        \includegraphics[width=\textwidth]{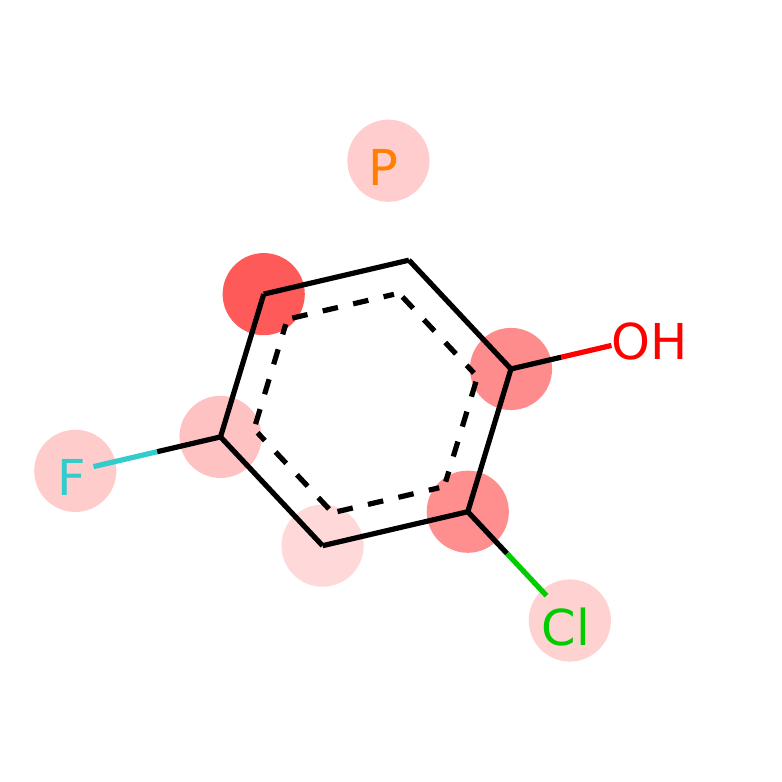}
        
    \end{subfigure}
    \hspace{0.1cm}
    \begin{subfigure}[b]{0.15\textwidth}
        \includegraphics[width=\textwidth]{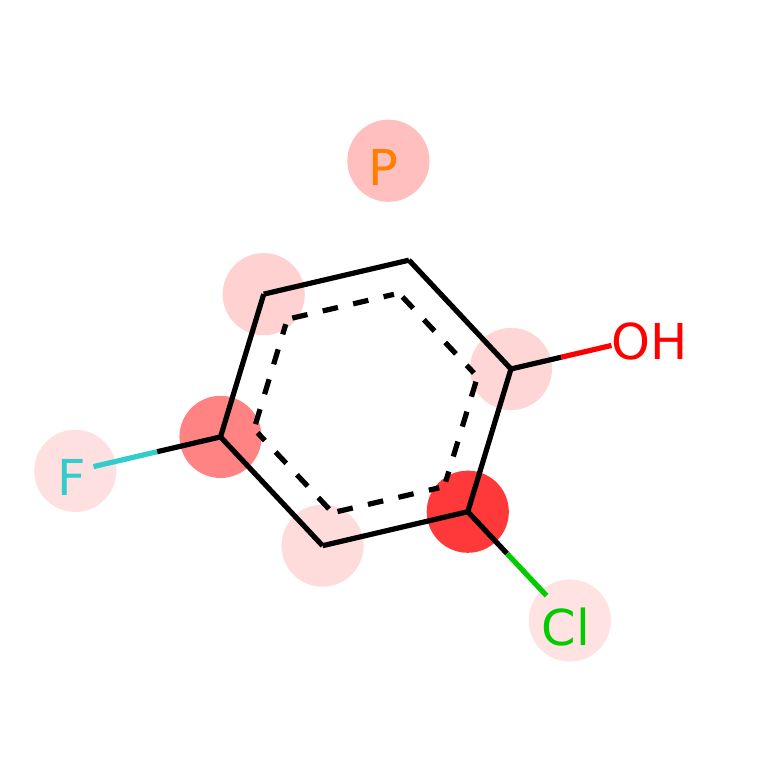}
        
    \end{subfigure}
    \hspace{0.1cm}
    \begin{subfigure}[b]{0.15\textwidth}
        \includegraphics[width=\textwidth]{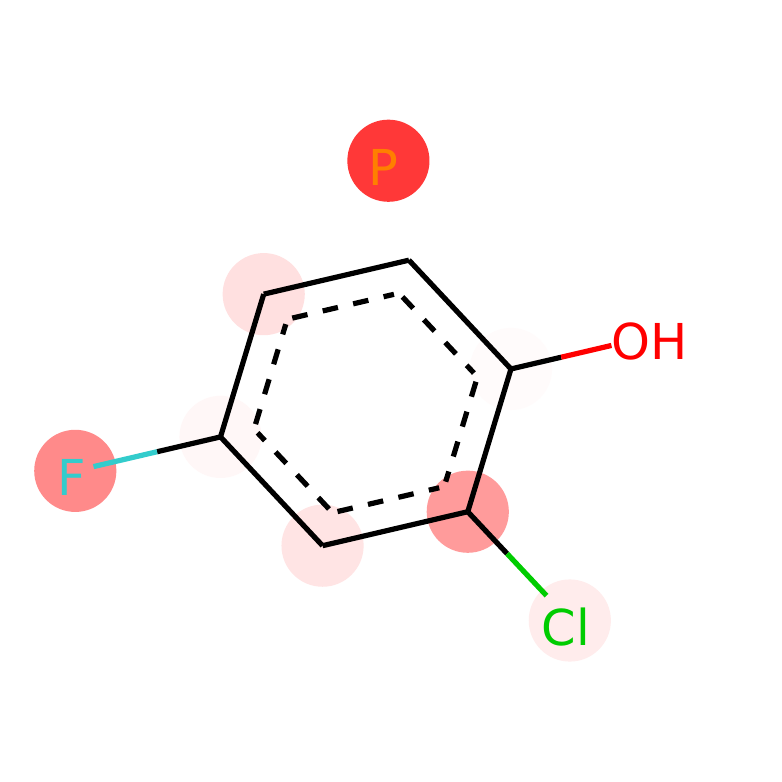}
        
    \end{subfigure}
    
  \label{fig:attplot}
  \caption{Attention probabilities: Every column corresponds to an attention head, and the rows are to the first, second and third layer respectively. The pool node depicted as a separate node.}
\end{figure}

\clearpage

\section*{Acknowledgment}
YM is funded by Research Council KU Leuven: C14/18/092 SymBioSys3; CELSA-HIDUCTION CELSA/17/032
Flemish Government:IWT: Exaptation, PhD grants
FWO 06260 (Iterative and multi-level methods for Bayesian multirelational factorization with features). This research received funding from the Flemish Government under the “Onderzoeksprogramma Artificiële Intelligentie (AI) Vlaanderen” program. EU: “MELLODDY” This project has received funding from the Innovative Medicines Initiative 2 Joint Undertaking under grant agreement No 831472. This Joint Undertaking receives support from the European Union’s Horizon 2020 research and innovation program and EFPIA. EDB is funded by a FWO-SB grant.

\ifCLASSOPTIONcaptionsoff
  \newpage
\fi



%
\bibliography{references}

\begin{thebibliography}{10}
\providecommand{\url}[1]{#1}
\csname url@samestyle\endcsname
\providecommand{\newblock}{\relax}
\providecommand{\bibinfo}[2]{#2}
\providecommand{\BIBentrySTDinterwordspacing}{\spaceskip=0pt\relax}
\providecommand{\BIBentryALTinterwordstretchfactor}{4}
\providecommand{\BIBentryALTinterwordspacing}{\spaceskip=\fontdimen2\font plus
\BIBentryALTinterwordstretchfactor\fontdimen3\font minus
  \fontdimen4\font\relax}
\providecommand{\BIBforeignlanguage}[2]{{%
\expandafter\ifx\csname l@#1\endcsname\relax
\typeout{** WARNING: IEEEtran.bst: No hyphenation pattern has been}%
\typeout{** loaded for the language `#1'. Using the pattern for}%
\typeout{** the default language instead.}%
\else
\language=\csname l@#1\endcsname
\fi
#2}}
\providecommand{\BIBdecl}{\relax}
\BIBdecl

\bibitem{duvenaud2015convolutional}
D.~K. Duvenaud, D.~Maclaurin, J.~Iparraguirre, R.~Bombarell, T.~Hirzel,
  A.~Aspuru-Guzik, and R.~P. Adams, ``Convolutional networks on graphs for
  learning molecular fingerprints,'' in \emph{Advances in neural information
  processing systems}, 2015, pp. 2224--2232.

\bibitem{kipf2016semi}
T.~N. Kipf and M.~Welling, ``Semi-supervised classification with graph
  convolutional networks,'' \emph{arXiv preprint arXiv:1609.02907}, 2016.

\bibitem{li2015gated}
Y.~Li, D.~Tarlow, M.~Brockschmidt, and R.~Zemel, ``Gated graph sequence neural
  networks,'' \emph{arXiv preprint arXiv:1511.05493}, 2015.

\bibitem{kearnes2016molecular}
S.~Kearnes, K.~McCloskey, M.~Berndl, V.~Pande, and P.~Riley, ``Molecular graph
  convolutions: moving beyond fingerprints,'' \emph{Journal of computer-aided
  molecular design}, vol.~30, no.~8, pp. 595--608, 2016.

\bibitem{velivckovic2017graph}
P.~Veli{\v{c}}kovi{\'c}, G.~Cucurull, A.~Casanova, A.~Romero, P.~Lio, and
  Y.~Bengio, ``Graph attention networks,'' \emph{arXiv preprint
  arXiv:1710.10903}, 2017.

\bibitem{yu2015MultiScaleContextAggregation}
\BIBentryALTinterwordspacing
F.~Yu and V.~Koltun, ``Multi-{Scale} {Context} {Aggregation} by {Dilated}
  {Convolutions},'' \emph{arXiv:1511.07122 [cs]}, Nov. 2015, arXiv: 1511.07122.
  [Online]. Available: \url{http://arxiv.org/abs/1511.07122}
\BIBentrySTDinterwordspacing

\bibitem{vaswani2017AttentionAllYou}
\BIBentryALTinterwordspacing
A.~Vaswani, N.~Shazeer, N.~Parmar, J.~Uszkoreit, L.~Jones, A.~N. Gomez,
  L.~Kaiser, and I.~Polosukhin, ``Attention is {All} you {Need},'' in
  \emph{Advances in {Neural} {Information} {Processing} {Systems} 30},
  I.~Guyon, U.~V. Luxburg, S.~Bengio, H.~Wallach, R.~Fergus, S.~Vishwanathan,
  and R.~Garnett, Eds.\hskip 1em plus 0.5em minus 0.4em\relax Curran
  Associates, Inc., 2017, pp. 5998--6008. [Online]. Available:
  \url{http://papers.nips.cc/paper/7181-attention-is-all-you-need.pdf}
\BIBentrySTDinterwordspacing

\bibitem{ba2016LayerNormalization}
\BIBentryALTinterwordspacing
J.~L. Ba, J.~R. Kiros, and G.~E. Hinton, ``Layer {Normalization},''
  \emph{arXiv:1607.06450 [cs, stat]}, Jul. 2016, arXiv: 1607.06450. [Online].
  Available: \url{http://arxiv.org/abs/1607.06450}
\BIBentrySTDinterwordspacing

\bibitem{he2016identity}
K.~He, X.~Zhang, S.~Ren, and J.~Sun, ``Identity mappings in deep residual
  networks,'' in \emph{European conference on computer vision}.\hskip 1em plus
  0.5em minus 0.4em\relax Springer, 2016, pp. 630--645.

\bibitem{xu2018powerful}
\BIBentryALTinterwordspacing
K.~Xu, W.~Hu, J.~Leskovec, and S.~Jegelka, ``How powerful are graph neural
  networks?'' in \emph{7th International Conference on Learning
  Representations, {ICLR} 2019, New Orleans, LA, USA, May 6-9, 2019}, 2019.
  [Online]. Available: \url{https://openreview.net/forum?id=ryGs6iA5Km}
\BIBentrySTDinterwordspacing

\bibitem{alzaga2008spectra}
A.~Alzaga, R.~Iglesias, and R.~Pignol, ``Spectra of symmetric powers of graphs
  and the weisfeiler-lehman refinements,'' \emph{arXiv preprint
  arXiv:0801.2322}, 2008.

\bibitem{scarselli2008graph}
F.~Scarselli, M.~Gori, A.~C. Tsoi, M.~Hagenbuchner, and G.~Monfardini, ``The
  graph neural network model,'' \emph{IEEE Transactions on Neural Networks},
  vol.~20, no.~1, pp. 61--80, 2008.

\bibitem{almeida1987learning}
L.~B. Almeida, ``A learning rule for asynchronous perceptrons with feedback in
  a combinatorial environment.'' in \emph{Proceedings, 1st First International
  Conference on Neural Networks}, vol.~2.\hskip 1em plus 0.5em minus
  0.4em\relax IEEE, 1987, pp. 609--618.

\bibitem{cho14gru}
\BIBentryALTinterwordspacing
K.~Cho, B.~van Merrienboer, {\c{C}}.~G{\"{u}}l{\c{c}}ehre, F.~Bougares,
  H.~Schwenk, and Y.~Bengio, ``Learning phrase representations using {RNN}
  encoder-decoder for statistical machine translation,'' \emph{CoRR}, vol.
  abs/1406.1078, 2014. [Online]. Available:
  \url{http://arxiv.org/abs/1406.1078}
\BIBentrySTDinterwordspacing

\bibitem{pearl2014probabilistic}
J.~Pearl, \emph{Probabilistic reasoning in intelligent systems: networks of
  plausible inference}.\hskip 1em plus 0.5em minus 0.4em\relax Elsevier, 2014.

\bibitem{dai2016discriminative}
H.~Dai, B.~Dai, and L.~Song, ``Discriminative embeddings of latent variable
  models for structured data,'' in \emph{International conference on machine
  learning}, 2016, pp. 2702--2711.

\bibitem{gilmer2017neural}
J.~Gilmer, S.~S. Schoenholz, P.~F. Riley, O.~Vinyals, and G.~E. Dahl, ``Neural
  message passing for quantum chemistry,'' in \emph{Proceedings of the 34th
  International Conference on Machine Learning-Volume 70}.\hskip 1em plus 0.5em
  minus 0.4em\relax JMLR. org, 2017, pp. 1263--1272.

\bibitem{hamilton2017inductive}
W.~Hamilton, Z.~Ying, and J.~Leskovec, ``Inductive representation learning on
  large graphs,'' in \emph{Advances in Neural Information Processing Systems},
  2017, pp. 1024--1034.

\bibitem{bruna2013spectral}
J.~Bruna, W.~Zaremba, A.~Szlam, and Y.~LeCun, ``Spectral networks and locally
  connected networks on graphs,'' \emph{arXiv preprint arXiv:1312.6203}, 2013.

\bibitem{defferrard2016convolutional}
M.~Defferrard, X.~Bresson, and P.~Vandergheynst, ``Convolutional neural
  networks on graphs with fast localized spectral filtering,'' in
  \emph{Advances in neural information processing systems}, 2016, pp.
  3844--3852.

\bibitem{li2018adaptive}
R.~Li, S.~Wang, F.~Zhu, and J.~Huang, ``Adaptive graph convolutional neural
  networks,'' in \emph{Thirty-Second AAAI Conference on Artificial
  Intelligence}, 2018.

\bibitem{bottou1997global}
L.~Bottou, Y.~Bengio, and Y.~Le~Cun, ``Global training of document processing
  systems using graph transformer networks,'' in \emph{Proceedings of IEEE
  Computer Society Conference on Computer Vision and Pattern
  Recognition}.\hskip 1em plus 0.5em minus 0.4em\relax IEEE, 1997, pp.
  489--494.

\bibitem{yun2019graph}
S.~Yun, M.~Jeong, R.~Kim, J.~Kang, and H.~J. Kim, ``Graph transformer
  networks,'' in \emph{Advances in Neural Information Processing Systems},
  2019, pp. 11\,960--11\,970.

\bibitem{li2019graph}
\BIBentryALTinterwordspacing
Y.~Li, X.~Liang, Z.~Hu, Y.~Chen, and E.~P. Xing, ``Graph transformer,'' 2019.
  [Online]. Available: \url{https://openreview.net/forum?id=HJei-2RcK7}
\BIBentrySTDinterwordspacing

\bibitem{maron2019provably}
H.~Maron, H.~Ben-Hamu, H.~Serviansky, and Y.~Lipman, ``Provably powerful graph
  networks,'' in \emph{Advances in Neural Information Processing Systems},
  2019, pp. 2153--2164.

\bibitem{jonas2019deep}
E.~Jonas, ``Deep imitation learning for molecular inverse problems,'' in
  \emph{Advances in Neural Information Processing Systems}, 2019, pp.
  4991--5001.

\bibitem{bento2014chembl}
A.~P. Bento, A.~Gaulton, A.~Hersey, L.~J. Bellis, J.~Chambers, M.~Davies, F.~A.
  Kr{\"u}ger, Y.~Light, L.~Mak, S.~McGlinchey \emph{et~al.}, ``The chembl
  bioactivity database: an update,'' \emph{Nucleic acids research}, vol.~42,
  no.~D1, pp. D1083--D1090, 2014.

\bibitem{wu2018moleculenet}
Z.~Wu, B.~Ramsundar, E.~N. Feinberg, J.~Gomes, C.~Geniesse, A.~S. Pappu,
  K.~Leswing, and V.~Pande, ``Moleculenet: a benchmark for molecular machine
  learning,'' \emph{Chemical science}, vol.~9, no.~2, pp. 513--530, 2018.

\bibitem{srivastava2014dropout}
N.~Srivastava, G.~Hinton, A.~Krizhevsky, I.~Sutskever, and R.~Salakhutdinov,
  ``Dropout: a simple way to prevent neural networks from overfitting,''
  \emph{The Journal of Machine Learning Research}, vol.~15, no.~1, pp.
  1929--1958, 2014.

\bibitem{kingma2014adam}
D.~P. Kingma and J.~Ba, ``Adam: A method for stochastic optimization,''
  \emph{arXiv preprint arXiv:1412.6980}, 2014.

\bibitem{jonas2019rapid}
E.~Jonas and S.~Kuhn, ``Rapid prediction of nmr spectral properties with
  quantified uncertainty,'' \emph{Journal of cheminformatics}, vol.~11, no.~1,
  pp. 1--7, 2019.

\bibitem{gunther1980nmr}
H.~Gunther, \emph{NMR spectroscopy: an introduction}.\hskip 1em plus 0.5em
  minus 0.4em\relax Wiley Chichester, 1980, vol.~81.

\bibitem{simm2018repurposing}
J.~Simm, G.~Klambauer, A.~Arany, M.~Steijaert, J.~K. Wegner, E.~Gustin,
  V.~Chupakhin, Y.~T. Chong, J.~Vialard, P.~Buijnsters \emph{et~al.},
  ``Repurposing high-throughput image assays enables biological activity
  prediction for drug discovery,'' \emph{Cell chemical biology}, vol.~25,
  no.~5, pp. 611--618, 2018.

\bibitem{rdkit}
\BIBentryALTinterwordspacing
G.~Landrum, B.~Kelley, P.~Tosco, sriniker, gedeck, N.~Schneider, R.~Vianello,
  A.~Dalke, A.~Savelyev, S.~Turk, M.~Swain, B.~Cole, A.~Vaucher,
  M.~W\'{o}jcikowski, A.~Pahl, JP, strets123, JLVarjo, P.~Fuller, D.~Gavid,
  N.~O'Boyle, D.~Cosgrove, G.~Sforna, M.~Nowotka, pzc, J.~van Santen, J.~H.
  Jensen, D.~Hall, D.~N, and P.~Avery, ``{RDKit}: Open-source
  cheminformatics,'' 2019. [Online]. Available: \url{http://www.rdkit.org}
\BIBentrySTDinterwordspacing

\bibitem{hartigan1975clustering}
J.~A. Hartigan, \emph{Clustering Algorithms}, 99th~ed.\hskip 1em plus 0.5em
  minus 0.4em\relax New York, NY, USA: John Wiley \& Sons, Inc., 1975.

\bibitem{doddrell1982distortionless}
D.~Doddrell, D.~Pegg, and M.~R. Bendall, ``Distortionless enhancement of nmr
  signals by polarization transfer,'' \emph{Journal of Magnetic Resonance
  (1969)}, vol.~48, no.~2, pp. 323--327, 1982.

\end{thebibliography}
\bibliographystyle{IEEEtran}

%

\end{document}